\documentclass[runningheads]{llncs}

\usepackage{amsfonts, amsmath, amsthm}
\usepackage[T1]{fontenc}
\usepackage{microtype}
\usepackage{graphicx}
\usepackage{subfigure}
\usepackage{booktabs} 
\usepackage{makecell}
\usepackage{graphicx}
\usepackage[lined,linesnumbered,ruled,commentsnumbered,resetcount,vlined]{algorithm2e}
\newcommand\SVDeq{\mathrel{\stackrel{\makebox[0pt]{\mbox{\normalfont\tiny SVD}}}{\,=\,}}}

\newcommand{\norm}[1]{\left\lVert#1\right\rVert}

\newcommand\QReq{\mathrel{\stackrel{\makebox[0pt]{\mbox{\normalfont\tiny QR}}}{\,=\,}}}

\newcommand{\startsquarepar}{%
	\par\begingroup \parfillskip 0pt \relax}
\newcommand{\stopsquarepar}{%
	\par\endgroup}

\begin{document}

\newtheorem*{Proposition_3_1}{Proposition 3.1: Error of \textsc{b-kfac} vs Error of (low projection error) \textsc{rs-kfac}}

\newtheorem*{Proposition_3_2}{Proposition 3.2: Pure \textsc{b-kfac} vs over-writing $\mathcal B_i = \tilde {\mathcal M}_{R;i,r}$ exactly once}
\newtheorem*{Proposition_4_1}{Proposition 4.1: Error of Doing nothing vs Error of B-updates}
\newtheorem*{Proposition_4_2}{Proposition 4.2: $\norm{E_j}_F$ ($j\geq 1$) Comparison for No-update vs for B-update}
\title{Brand New K-FACs: Speeding up K-FAC with Online Decomposition Updates}
\titlerunning{Brand New KFACs}
%
%
\author{Constantin Octavian Puiu\orcidID{0000-0002-1724-4533}}
\institute{University of Oxford, Mathematical Institute,\\
	\email{constantin.puiu@maths.ox.ac.uk}\\
}

\maketitle              

\authorrunning{C. O. Puiu}
%

\vskip 0.3in



\begin{abstract}
	\textsc{k-fac} (\cite{KFAC}, \cite{convolutional_KFAC}) is a tractable implementation of Natural Gradient for Deep Learning, whose bottleneck is computing the inverses of the so-called ``Kronecker-Factors''. \textsc{rs-kfac} (\cite{randomized_KFACS}) is a \textsc{k-fac} improvement which provides a cheap way of estimating the K-factors inverses. In particular, it reduces the cubic scaling (in layer width) of standard K-FAC down to quadratic. In this paper, we exploit the exponential-average construction paradigm of K-Factors, and use  online-NLA techniques (\cite{Brand2006}) to propose an even cheaper (but less accurate) way of estimating the K-factors inverses for FC layers. In particular, we propose a K-factor inverse update which scales linearly in layer size. We also propose an inverse application procedure which scales linearly as well (the one of \textsc{k-fac} scales cubically and the one of \textsc{rs-kfac} scales quadratically). Overall, our proposed algorithm gives a \textsc{k-fac} implementation whose preconditioning part scales linearly in layer size (compare to cubic for \textsc{k-fac} and quadratic for \textsc{rs-kfac}). Importantly however, this update is only applicable in some circumstances, unlike the \textsc{rs-kfac} approach \cite{randomized_KFACS}. 
	
	The inverse updates proposed here can be combined with \textsc{rs-kfac} updates to give different algorithms. Numerical results show \textsc{rs-kfac}'s (\cite{randomized_KFACS})  inversion error can be reduced with minimal time overhead by adding our proposed update to it. Based on the proposed procedure, a correction to it, and \textsc{rs-kfac}, we propose three practical algorithms for optimizing generic Deep Neural Nets. Numerical results show that two of these outperform \textsc{rs-kfac} (\cite{randomized_KFACS}) for any target test accuracy on CIFAR10 classification with a slightly modified version of VGG16\_bn. Our proposed algorithms achieve 91$\%$ test accuracy faster than \textsc{seng} (\cite{SENG}) but underperform it for higher test-accuracy.
	
\end{abstract}
\section{Introduction}
	The desirable properties (\cite{New_insights_and_perspectives}) of Natural Gradient (NG; \cite{AMARI}) has determined research in optimization for Deep Learning (DL) to lately focus on developing (and improving) tractable NG implementations. K-FAC (\cite{KFAC}, \cite{convolutional_KFAC}) is such a tractable implementation of NG for DL which makes substantial progress per epoch, but requires computing the inverses of the so-called ``Kronecker-Factors'' (K-Factors). While tractable, computing these inverses can become very slow for wide nets \cite{SENG}. By noting that the exponential-average (EA) construction paradigm of the K-Factors leads to eigen-spectrum decay, a way to significantly speed up the K-factors inversion using randomized linear algebra was proposed in \cite{randomized_KFACS}. 
	
	In this paper, we exploit the EA construction paradigm of K-Factors in a  different fashion, by using online-NLA techniques (\cite{Brand2006}), and propose an even cheaper\footnote{Than the ``inversion'' procedure proposed in \cite{randomized_KFACS}.} (but less accurate) way of estimating the K-factors inverses for FC layers. 
	Our \textbf{contributions} are as follows:
\begin{enumerate}
	\item \textbf{Linear Time Inverse Computation.} Proposing a new, cheaper way of performing the ``inverse computation'' in \textsc{k-fac} (\cite{KFAC}) for FC layers, exploiting the online construction paradigm of exponentially-averaged K-Factors (and the eigenspectrum decay thereof). W.r.t.\ the very recently proposed randomized approach of \textsc{rs-kfac} (in \cite{RSVD_paper}), our proposed update is faster, but provides lower ``inversion'' accuracy\footnote{However, we can trade-off speed to gain accuracy by increasing update frequency.}. This computation is linear in layer size (compare to quadratic for randomized K-FACs \cite{randomized_KFACS} and cubic for standard K-FAC \cite{KFAC}, \cite{convolutional_KFAC}). See \textit{Section 3}.
	
	\item \textbf{Linear Time Inverse Application.} Proposing a way to apply the (proposed, low-rank) inverse representation of K-factors onto the Gradient whose time scales linearly in layer size (compare to quadratic for randomized K-FACs \cite{randomized_KFACS} and cubic for standard K-FAC \cite{KFAC}, \cite{convolutional_KFAC}). We only propose this, but the numerical results herein do not use it yet (see \textit{Section 5}). Implementing it is future work.
	
	\item Simple theoretical results showing that under worst-case scenarios our proposed update to K-Factors inverses is strictly better than no update. This result is valuable in the context of the update being very cheap. See \textit{Sections 3} and \textit{4}.
	
	\item Numerical results showing our proposed online update (when introduced on top of the existing updates in a given \textsc{rs-kfac} algorithm) can significantly improve the error in K-Factors inverse with minimal computation time overhead. See \textit{Section 4}.
	
	\item 3 practical algorithms (\textit{``Brand New K-FACs''}) using our proposed update, possibly combined with \textsc{rs-kfac} updates, and a ``correction'' we introduce. See \textit{Section 3}.

	\item Numerical results (for a particular case-study) showing that two of the \textit{Brand New K-FACs} (\textsc{b-kfac}, \textsc{b-kfac-c}) outperform \textsc{rs-kfac} for all the considered target test accuracies, while the other one (\textsc{b-r-kfac}) does so (only) for high target test accuracy. See \textit{Section 6}.
	
	\item Numerical results (for a particular case-study) showing \textsc{b-r-kfac} is better than \textsc{k-fac} (\cite{KFAC}) for 3/4 error metrics, while being almost on par for the $4^{\text{th}}$ metric and $3\times$ cheaper. See \textit{Section 6}.

\end{enumerate}

\subsubsection{Related Work} Puiu (2022, \cite{randomized_KFACS}) proposes to speed up K-Factors inversion using randomized NLA. Tang et.\ al.\ (2021, \cite{low_rank_Kfac_SKFAC}) proposes to construct a more efficient inversion of the regularized low-rank K-factors by using the Woodbury formula to express the K-Factors inverses in terms of $A_i^{(l)}$ and $G_i^{(l)}$ (see \textit{Section 2.2}). In contrast with our proposed K-Factors inverse update, none of the two approaches employs online NLA or Brand's algorithm (\cite{Brand2006}). Osawa et.\ al.\ (2020, \cite{Kazuki_SPNGD}) presents some ideas to speed-up \textsc{k-fac}, but they are completely different to ours.
\section{Preliminaries}

\subsubsection{Neural Networks and Supervised Learning} We focus on the case of supervised learning for simplicity, but our proposed update can be used whenever K-FAC can be applied (but only improves computational time for FC layers computation). 

We have a dataset $\mathcal D = \{(x_i,y_i)\}_{i=1,..., N_{\mathcal D}}$ of $N_{\mathcal D}$ input-target pairs $\{x_i,y_i\}$. Let us consider a DNN $h_{\theta}(\cdot)$ with $n_L$ layers, where $\theta$ are the aggregated network parameters. We denote the predictive distribution of the network (over labels - e.g.\ over classes) by $p(y|h_\theta(x))$ (shorthand notation $p_\theta (y|x)$). Note that this is parameterized by $h_\theta(x)$. Our learning problem is 
\begin{equation}
\min_{\theta} f(\theta) := \frac{1}{N_{\mathcal D}}\sum_{(x_i,y_i)\in\mathcal D}\big(- \log p (y_i|h_\theta(x_i))\big).
\label{eqn_optimization_problem}
\end{equation}
\noindent We let $g_k:=\nabla_\theta f(\theta_k)$, and note that we can express $g_k = [g_k^{(1)}, ..., g_k^{(n_L)}]$, where $g_k^{(l)}$ is the gradient of parameters in layer $l$. We will use a superscript to refer to the \textit{layer} index and a subscript to refer to the \textit{optimization iteration} index. 

\subsection{Fisher, Natural Gradient Descent and K-FAC}
For our purposes, the Fisher Information matrix is defined as
\begin{equation}
F_k := F(\theta_k) := \mathbb E_{\substack{ x\sim \mathcal D \\ y\sim p_\theta(y|x)}}\biggl[\nabla_\theta \log p_\theta(y|x) \nabla_\theta \log p_\theta(y|x)^T\biggr].
\label{eqn_FIM}
\end{equation}
The Natural gradient descent step is defined as
\begin{equation}
s^{(\text{NGD})} = - F_k^{-1}g_k.
\end{equation}

Typically, the number of parameters $|\theta|$ is very large. In this case, storing and linear-solving with the Fisher is infeasible. \textsc{k-fac} is an algorithm that offers a solution to this issue by approximating $F_k$ as a block-diagonal matrix, where each block is represented as the Kronecker factor of two smaller matrices \cite{KFAC}. We have
\begin{equation}
F_k\approx F_k^{(\text{KFAC})} := \text{blockdiag}\big(\{\mathcal A^{(l)}_k \otimes \Gamma^{(l)}_k \}_{l=1,...,n_L}\big),
\label{KFAC_matrix_defn}
\end{equation}
where $\mathcal A^{(l)}_k:= A_k^{(l)}[A_k^{(l)}]^T$ and $\Gamma^{(l)}_k:= G_k^{(l)}[G_k^{(l)}]^T$ are the \textit{forward K-factor} and \textit{backward K-factor} respectively (of layer $l$ at iteration $k$) \cite{KFAC}. The exact K-Factors definition depends on the layer type (see \cite{KFAC} for FC layers, \cite{convolutional_KFAC} for Conv layers). For our purpose, it is sufficient to state that $A_k^{(l)}\in\mathbb R^{d^{(l)}_A \times n^{(l)}_M}$ and $G_k^{(l)}\in\mathbb R^{d^{(l)}_{\Gamma}\times n^{(l)}_M}$, with $n^{(l)}_M \propto n_{\text{BS}}$ for Convolutonal layers and $n^{(l)}_M = n_{\text{BS}}$ for FC layers, where $n_{\text{BS}}$ is the batch size. In \textsc{k-fac}, $(F_k^{(\text{KFAC})})^{-1}g_k$ is computed by first performing an eigenvalue decomposition (EVD) of the Kronecker factors (${\mathcal A}^{(l)}_k$ and $\Gamma^{(l)}_k$), and then noting that $(\mathcal A^{(l)}_k \otimes \Gamma^{(l)}_k)^{-1}g_k^{(l)} = \text{vec}\big([\Gamma^{(l)}_k]^{-1} \text{Mat}(g_k^{(l)}) [\mathcal A^{(l)}_k]^{-1}\big)$, where $\text{vec}(\cdot)$ is the matrix vectorization operation and $\text{Mat}(\cdot)$ is its inverse \cite{KFAC}.
\vspace{-1.5ex}

\subsubsection{K-FAC in practice}  Let $\mathbb I_{\{\cdot\}}$ be the indicator function and $\kappa(i) := 1 - \rho \mathbb I_{\{i>0\}}$. In practice, an exponential average (EA) is held for the K-factors. Thus, we use $\bar{\mathcal A}_k^{(l)}$ and $\bar{\Gamma}_k^{(l)}$ instead of ${\mathcal A}_k^{(l)}$ and ${\Gamma}_k^{(l)}$ in the discussion above, where
\vspace{-2.5ex}

\begin{equation}
\bar {\mathcal A}_k^{(l)} := \sum_{i=0}^k\kappa(i)\rho^{k-i}A_i^{(l)}[A_i^{(l)}]^T,\,\,\,\, \bar \Gamma_k^{(l)} := \sum_{i=0}^k \kappa(i)\rho^{k-i}G_i^{(l)}[G_i^{(l)}].
\label{eqn_EA_kfac}
\end{equation}
\subsection{Randomized K-FACs} 
\begin{algorithm}[H]
	\caption{\textsc{r-kfac} (\textsc{rs-kfac}, \cite{randomized_KFACS})}
	\label{R_KFAC_algorithm}
	\For{$k=0,1,2,....$}{
		
		Choose batch $\mathcal B_k \subset \mathcal D$
		
		\For(\tcp*[h]{Perform forward pass}){$l=0,1,...,N_L$}
		{
			Get $a^{(l)}_k$ and $A^{(l)}_k$
			
			$\bar {\mathcal A}^{(l)}_k \leftarrow \rho \bar A^{(l)}_{k-1} + (1-\rho) A^{(l)}_k [A^{(l)}_k]^T$ \tcp{Update fwd.\ statistics}
			
		}
		Get $\tilde f(\theta_k)$ \tcp*{From $a^{(l)}_k$}
		
		\For(\tcp*[h]{Perform backward pass}){$l=N_L,N_{L-1},...,1$}
		{
			Get $g^{(l)}_k$ and $G^{(l)}_k$
			
			$\bar \Gamma^{(l)}_k \leftarrow \rho \bar \Gamma^{(l)}_{k-1} + (1-\rho) G^{(l)}_k [G^{(l)}_k]^T$ \tcp{Update bwd.\ statistics}
		}
		Get gradient $g_k = \big[\big(g^{(1)}_k\big)^T,... \big(g^{(N_L)}_k\big)^T\big]^T$
		
		\For(\tcp*[h]{Compute \textsc{k-fac} step:}){$l=0,1,...,N_L$}
		{
			\tcp{Get RSVD of $\bar {\mathcal A}^{(l)}_k$ and $\bar \Gamma^{(l)}_k$ for inverse application}
			$\tilde U^{(l)}_{A,k}\tilde D^{(l)}_{A,k} (\tilde V^{(l)}_{A,k})^T = \text{RSVD}(\bar {\mathcal A}^{(l)}_k)$;	$\tilde U_{\Gamma,k}^{(l)}\tilde D_{\Gamma,k}^{(l)} (\tilde V^{(l)}_{\Gamma,k})^T = \text{RSVD}(\bar \Gamma^{(l)}_k)$
			
			\tcp{Use RSVD to approx.\ apply inverse of K-FAC matrices}
			
			$J^{(l)}_k = \text{Mat}(g_k^{(l)})$
			
			$M^{(l)}_k = J^{(l)}_k \tilde V^{(l)}_{A, k}\big[( \tilde D^{(l)}_{A,k} + \lambda_{k,l}^{(\mathcal A)} I)^{-1} - \frac{1}{\lambda_{k,l}^{(\mathcal A)}}I\big] (\tilde V^{(l)}_{A,k})^T+ \frac{1}{\lambda_{k,l}^{(\mathcal A)}}J^{(l)}_k$
			
			$S^{(l)}_k = \tilde V^{(l)}_{\Gamma, k}\biggl[( \tilde D^{(l)}_{\Gamma,k} + \lambda_{k,l}^{(\Gamma)} I)^{-1} - \frac{1}{\lambda_{k,l}^{(\Gamma)}}I\biggr] (\tilde V^{(l)}_{\Gamma,k})^T M^{(l)}_k + \frac{1}{\lambda_{k,l}^{(\Gamma)}}M^{(l)}_k $
			
			$s^{(l)}_k = \text{vec}(S^{(l)}_k)$
		}
		
		$\theta_{k+1} = \theta_k - \alpha_k[(s_k^{(1)})^T,...,(s_k^{(N_L)})^T]^T$ 	\tcp{Take K-FAC step}
		
	}	
\end{algorithm}
The approach in \textsc{k-fac} is \textit{relatively} efficient, since the dimensions of the K-factors ($\bar{\mathcal A}^{(l)}_k$ and $\bar \Gamma^{(l)}_k$) is smaller than the dimension of the blocks which would have to be inverted in the absence of the Kronecker factorization \cite{KFAC}. However, these K-factors sometimes get large enough that the eigen-decomposition is very slow. A solution to this problem which exploits the rapid decay of the K-Factors eigenspectrum is proposed in \cite{randomized_KFACS}. Two algorithms which substantially speed up \textsc{k-fac} are proposed: \textsc{rs-kfac} and \textsc{sre-kfac} (generically called ``\textit{Randomized K-FACs}'') \cite{randomized_KFACS}. These algorithms essentially replace the eigen-decomposition of \textsc{k-fac} with \textit{randomized\footnote{Details about randomized SVD/EVD can be found in \cite{RSVD_paper}. Summary of these in \cite{randomized_KFACS}.} SVD} (in the case of \textsc{rs-kfac}) or \textit{randomized eigenvalue decomposition} (in the case of \textsc{sre-kfac}) \cite{randomized_KFACS}. 

Over-all, \textsc{r-kfac}'s time cost scales like $\mathcal O(d_M^2(r+r_o))$ (when setting $d_M=d_{\mathcal A}^{(l)}=d_{\Gamma}^{(l)}$ and $r = r_{\mathcal A}^{(l)} = r_{\Gamma}^{(l)}$, $\forall l$, for simplicity of exposition). Note that $r_{\mathcal A}^{(l)}$ and $r_{\Gamma}^{(l)}$ are the target-ranks of our low-rank representation for K-factors $\bar{\mathcal A}_k^{(l)}$ and $\bar{\Gamma}_k^{(l)}$ respectively, and $r_o$ is the \textsc{rsvd} oversampling parameter.

We will construct our discussion starting from these \textit{Randomized K-FACs}. We only present \textsc{rs-kfac} (the most successful in \cite{randomized_KFACS}) in \textit{Algorithm 1} for convenience. For convenience, we will from now on use ``\textsc{r-kfac}'' to denote the \textsc{rs-kfac} in \cite{randomized_KFACS}.

\vspace{+2ex}
\noindent \textbf{RSVD and EA update Frequencies Note: }In practice we perform \textit{lines 5} and \textit{9} only once every $T_{\text{updt}}$ iterations, and \textit{line 13} only once  every $T_{\text{inv}}$ iterations. We omitted the corresponding \textit{if} statements in \textit{Algorithm 1} for convenience.


\subsection{Brand's Algorithm 2006}
We now look at an algorithm which allows us to cheaply update the thin-SVD of a low-rank matrix when the original matrix is updated through a low-rank addition. We will refer to this as the Brand algorithm\footnote{Word of warning: there exist other algorithms by Brand M.} (proposed in \cite{Brand2006}, 2006). Consider the low-rank matrix $X\in\mathbb R^{m\times d}$, with rank $r<\min(m,d)$ and its thin SVD
\begin{equation}
X \SVDeq U D V^T,
\end{equation}
where $U\in\mathbb R^{m\times r}$, $V\in\mathbb R^{d\times r}$ are orthonormal matrices and $D\in\mathbb R^{r\times r}$ is diagonal. Now, suppose we want to compute the SVD of $\hat X := X + AB^T$, where $A\in\mathbb R^{m\times n}$ and $B\in\mathbb R^{d\times n}$ with $n$ s.t.\ $r+n <\min(m,d)$. Brand's Algorithm (exactly) computes this SVD cheaper than performing the SVD of $\hat X$ from scratch, by exploiting the available SVD of $X$ \cite{Brand2006}. To do so, it uses the identity \cite{Brand2006}
\begin{equation}
\hat X =  \begin{bmatrix}
U & Q_A
\end{bmatrix}  M_S
\begin{bmatrix}
V & Q_B
\end{bmatrix}^T\,\,\text{with}\,\,\,M_S := \begin{bmatrix}
I & U^TA\\
0 & R_A
\end{bmatrix}
\begin{bmatrix}
D & 0\\
0 & I
\end{bmatrix}
\begin{bmatrix}
I & V^TB\\
0 & R_B
\end{bmatrix}^T,
\label{eqn_MS_brand_algo}
\end{equation}
\noindent where $Q_A R_A \QReq (I-UU^T)A$ and $Q_B R_B \QReq (I-VV^T)B$ are the QR decompositions\footnote{Any decomposition where $Q_A$ and $Q_B$ are orthonormal matrices would work, but we pin it down to QR for simplicity. See Brand's paper \cite{Brand2006}.} of matrices $(A-UU^TA)$ and $(B-VV^TB)$ respectively. Now, we only need to perform the SVD of the \textit{small} matrix $M_S\in \mathbb R^{(r+n)\times (r+n)}$. We can then use the SVD of $M_S$ to obtain the SVD of $\hat X$ (as $U$, $V$, $Q_A$ and $Q_B$ are orthonormal and $U^TQ_A = V^TQ_B = 0$)\footnote{The reader is referred to the original paper for details \cite{Brand2006}.}. Brand's algorithm is shown below. 

The time complexity of \textit{Algorithm 2} is $\mathcal O((r+n)^4 + (m + d)(r+n)^2)$. This is better than performing \textsc{rsvd} (\cite{RSVD_paper}) on $\hat X$ with target rank $r+n$, which is $\mathcal O(mn(r+n+r_o)^2)$ (typically $r_o\approx 10$). Note that Brand's algorithm gives the exact SVD. The \textsc{rsvd} would also be (almost) exact when the target rank is $r+n$.

\begin{algorithm}[H]
	\caption{Brand's algorithm \cite{Brand2006}}
	\textbf{Input:} $U\in\mathbb R^{m\times r}$, $V\in\mathbb R^{d\times r}$, $D\in\mathbb R^{r\times r}$, $A\in\mathbb R^{m\times n}$ and $B\in\mathbb R^{d\times n}$ \tcp{with $(r+n)<\min(m,d)$} 
	
	\textbf{Output:} SVD of $\hat X : = X + AB^T = UDV^T + AB^T$
	
	Compute $U^TA$ and $V^TB$ \tcp*{$\mathcal O\big((m+d)rn\big)$ flops}
	
	$A_{\perp} := A-UU^TA$; $B_{\perp} :=  B-VV^TB$\tcp*{$\mathcal O\big((m+d)rn\big)$ flops}

	$Q_A R_A = \text{QRdec}(A_{\perp})$; $Q_B R_B = \text{QRdec}(B_{\perp})$ \tcp*{$\mathcal O\big((m+d)n^2\big)$ flops}

	Assemble $M_S$ as in (\ref{eqn_MS_brand_algo}) \tcp*{$\mathcal O\big((r+n)^3\big)$ flops}
	
	$U_M D_M V_M^T = \text{SVD}(M_S)$ \tcp*{$\mathcal O((r+n)^4)$ flops}
	
	Compute $U_{\hat X} = \begin{bmatrix}
	U & Q_A
	\end{bmatrix} U_M$ \tcp*{$\mathcal O\big(d(r+n)^2\big)$ flops}
	
	Compute  $V_{\hat X} = \begin{bmatrix}
	V & Q_B
	\end{bmatrix} V_M$ \tcp*{$\mathcal O\big(m(r+n)^2\big)$ flops}
	
	Set $D_{\hat X} = D_M$
	
	\textbf{Return} SVD of $\hat X$: $U_{\hat X} D_{\hat X} V_{\hat X}^T  \SVDeq \hat X$
\end{algorithm}

\subsubsection{Brand's Algorithm for Symmetric $X$ with Symmetric Update}
In our case, we only care about the case when $X\in\mathbb R^{d\times d}$ is square, symmetric and positive semi-definite: with SVD $X\SVDeq UDU^T$ and $A=B$. In this case, the SVD and EVD of $\hat X$ will be the same (and also for $X$), and we can also spare some computation. The Symmetric Brand's algorithm is shown in \textit{Algorithm 3} (this is our own trivial adaptation after Brand's Algorithm to use the symmetry).

Note that $M_S$ will be symmetric in this case. Furthermore the eigenvalues of $M_S$ will be the same as the eigenvalues of $\hat X$, which are nonnegative. Thus, $M_S$'s SVD and EVD are the same. Thus, we can compute $U_M$ and $D_M$ in practice by using a \textit{symmetric eigenvalue decomposition} algorithm of the small matrix $M_S$. 

\begin{algorithm}[H]
	\caption{Symmetric Brand's algorithm}
	\textbf{Input:} $U\in\mathbb R^{d\times r}$, $D\in\mathbb R^{r\times r}$, $A\in\mathbb R^{d\times n}$ \tcp{with $(r+n)<d$} 
	
	\textbf{Output:} SVD of $\hat X : = X + AA^T = UDU^T + AA^T$
	
	Compute $U^TZ$; Compute $Z_{\perp} := A-UU^TA$ \tcp*{$\mathcal O\big(drn\big)$ flops}
	
	$Q_A R_A = \text{QR\_decomp}(A_{\perp})$ \tcp*{$\mathcal O(dn^2)$ flops}

	Assemble $M_S$ as in (\ref{eqn_MS_brand_algo}) with $B\leftarrow A$, and $V\leftarrow U$ \tcp*{$\mathcal O\big((r+n)^3\big)$ flops}
	
	$U_M D_M U_M^T = \text{EVD}(M_S)$ \tcp*{$\mathcal O((r+n)^4)$ flops}
	
	Compute $U_{\hat X} = \begin{bmatrix}
	U & Q_A
	\end{bmatrix} U_M$; Set $D_{\hat X} = D_M$ \tcp*{$\mathcal O\big(d(r+n)^2\big)$ flops}

	\textbf{Return} (Exact) SVD of $\hat X$: $U_{\hat X} D_{\hat X} U_{\hat X}^T  \SVDeq \hat X$
\end{algorithm}

\noindent The total complexity of \textit{Algorithm 3} is $\mathcal O((r+n)^4 + d(r+n)^2)$. This is better than the complexity of directly performing \textsc{srevd}\footnote{Symmetric variant of \textsc{rsvd}, see \cite{randomized_KFACS}.} on $\hat X$, which is $\mathcal O (d^2(r+n+r_o))$ (for $r_o\approx10$), especially when $r+n \ll d$ (the case we will fall into in practice, at least for some K-factors). However, note the computational saving w.r.t. non-symmetric Brand's Algorithm is modest.

\subsubsection{Practical Considerations}
We have seen that using symmetric Brand's algorithm to adjust for a low-rank update is faster than performing the \textsc{srevd}, while giving the exact same result\footnote{\textsc{srevd} will (almost) give the exact EVD since the target rank is the true rank. We will need to use an ``oversampling" parameter $r_o\approx10$ say for this to happen though - which will not modify the complexity substantially. Brand's Algorithm is exact.}. However, to use Brand's algorithm we had to have the EVD of $X$ - which in principle requires further computation. We will see that in our case, because we work in an ``online'' setting, we can actually obtain an approximate EVD of $X$ for free. Thus, we can use Brand's algorithm to obtain further speed-ups when compared to merely using \textsc{srevd} (or \textsc{rsvd}), but at the expense of some accuracy (we will use an approximate EVD of ``$X$'').



\section{Linear Time (in Layer Size) EA K-Factors Inversion }
Consider the \textsc{r-kfac} algorithm\footnote{The discussion in this paragraph also applies to \textsc{sre-kfac}.} (\textit{Algorithm 1}). The key inefficiency of \textsc{r-kfac}, is that each time we compute an RSVD, we do so ``from scratch'', not using any of the previous RSVDs. Since we are always interested in the RSVD of a matrix which differs from a previous one (that we have the RSVD of) only through a low-rank update, further speed-ups can be obtained here. We now propose a way of obtaining such speed-ups by using the online algorithms presented in \textit{Section 2.3}. Doing so causes a further accuracy reduction in obtaining the ``inverses'' of K-Factors\footnote{In addition to the one introduced by using the \textsc{rsvd} in \textsc{r-kfac} instead of the \textsc{evd} as in \textsc{k-fac}.}, but this may be improved as described in \textit{Section 3.3}.

\subsection{Brand K-FAC (B-KFAC)}

The idea behind our approach in \textit{``Brand K-FAC''} is simple. Instead of performing an \textsc{rsvd} of $\bar {\mathcal A}^{(l)}_k$ and $\bar \Gamma^{(l)}_k$ at each step (as in \textsc{r-kfac}), we use  Brand's algorithm to update the previously held low-rank representation of the K-Factors (the $\tilde U$'s and $\tilde D's$) based on the incoming low-rank updates $(1-\rho) A^{(l)}_k (A^{(l)}_k)^T$ and $(1-\rho) \Gamma^{(l)}_k (\Gamma^{(l)}_k)^T$. Thus, we \textit{directly apply} Symmetric Brand's algorithm  to estimate a low-rank \textsc{svd} representation of $\bar A_{k+1}$ by replacing $U$, $D$ and $A$ in \textit{Algorithm 3} by $\tilde U^{(l)}_{A,k}$, $\rho\tilde D^{(l)}_{A,k}$ and  $\sqrt{1-\rho}  A^{(l)}_k$ respectively. We also perform an analogous replacement for $\bar \Gamma$. Importantly, we start our $\tilde U_{\cdot,0}$ and $\tilde D_{\cdot,0}$ (at $k=0$) from an RSVD in practice. The implementation is shown in \textit{Algorithm 4}.

In practice we perform \textit{lines 2-7} only once in $T_{\text{Brand}}$ steps.

\vspace{+2ex}
\noindent \textbf{Controlling the size of $\tilde U^{(l)}_{(\cdot),k}$'s and $\tilde D^{(l)}_{(\cdot),k}$'s:}  Each application of Brand's algorithm increases the size of carried matrices. To avoid indefinite size increase, we truncate $\tilde U^{(l)}_{A,k-1}\tilde D^{(l)}_{A,k-1}[\tilde U^{(l)}_{A,k-1}]^T$ to rank $r$ just before applying the Brand update (and similarly for $\Gamma$-related quantities). In other words, we enforce $\tilde U^{(l)}_{A,k-1} \in \mathbb R^{d_{\mathcal A}^{(l)}\times r}$, $\tilde U^{(l)}_{\Gamma,k-1} \in \mathbb R^{d_{\Gamma}^{(l)}\times r}$, $\tilde D^{(l)}_{A,k-1}, \tilde D^{(l)}_{\Gamma,k-1} \in \mathbb R^{r\times r}$ by retaining only the first $r$ modes just before \textit{lines 5-7} of \textit{Algorithm 4}. Note that by truncating just before applying Brand's algorithm, we use the $r+n_M^{(l)}$ rank approximation when applying our K-factors inverse. 

\begin{algorithm}[H]
	\caption{Brand \textsc{k-fac} (\textsc{b-kfac})}
	Replace \textit{lines 12 - 13 in Algorithm 1} with:

	\tcp{Truncate to rank $r$: maintain matrices sizes}	
	$\tilde U^{(l)}_{A,k-1}\leftarrow \tilde U^{(l)}_{A,k-1}[:,:r]$, $\tilde D^{(l)}_{A,k-1}\leftarrow\tilde D^{(l)}_{A,k-1}[:r,:r]$
	
	$U^{(l)}_{\Gamma,k-1}\leftarrow \tilde U^{(l)}_{\Gamma,k-1}[:,:r]$, $\tilde D^{(l)}_{\Gamma,k-1}\leftarrow \tilde D^{(l)}_{\Gamma,k-1}[:r,:r]$ 
	
	\tcp{Use Symmetric Brand's low-rank update (\textit{Algorithm\ 3})}
			$\tilde U^{(l)}_{A,k},\,\tilde D^{(l)}_{A,k} = \text{Symmetric\_Brand}(\tilde U^{(l)}_{A,k-1}, \rho \tilde D^{(l)}_{A,k-1}, \sqrt{1-\rho}\, A_k^{(l)})$ 
			
			$\tilde U^{(l)}_{\Gamma,k},\,\tilde D^{(l)}_{\Gamma,k} = \text{Symmetric\_Brand}(\tilde U^{(l)}_{\Gamma,k-1}, \rho \tilde D^{(l)}_{\Gamma,k-1}, \sqrt{1-\rho}\, G_k^{(l)})$ 

\end{algorithm}
Note that the complexity of obtaining our inverse representation is now $\mathcal O((r+n_{M}^{(l)})^4 + d_M(r+n_{M}^{(l)})^2)$ (when setting $d_M=d_{\mathcal A}^{(l)}=d_{\Gamma}^{(l)}$ and $r = r_{\mathcal A}^{(l)} = r_{\Gamma}^{(l)}$, $\forall l$, for simplicity of exposition). Compared to $\mathcal O(d_M^2(r+r_o))$ for \textsc{r-kfac} or $\mathcal O(d_M^3)$ for standard \textsc{k-fac} \cite{KFAC}, this is much better when $r + n_M\ll d_M$, in which case the over-all coplexity becomes linear in $d_M$: $\mathcal O(d_M(r+n_{M}^{(l)})^2)$. We shall see in \textit{Section 3.5} that $r + n_M\ll d_M$ typically holds for FC layers.

\vspace{+2ex}

\textbf{Error} \textbf{Comments:} Brand's algorithm is exact, but the truncations introduce an error in each of our low-rank K-Factors representations, at each $k$.

\subsection{Mathematically Comparing B-KFAC and R-KFAC Processes}
To better understand the connections and differences between \textsc{b-kfac} and \textsc{r-kfac} let us consider how the K-factor estimate (which is used to obtain the inverse) is constructed in both cases. Consider an arbitrary EA K-Factor $\mathcal M_k$ (may be either $\bar{\mathcal A}_k^{(l)}$ or $\bar{\Gamma}_k^{(l)}$ for any $l$) where we have incoming (random) updates $M_kM_k^T$ with $M_k\in\mathbb R^{d\times n_{\text{BS}}}$ at iteration $i$. This follows the process
\begin{equation}
\begin{split}
\mathcal M_0 = M_0M_0^T,\,\,\,
\mathcal M_{j}= \rho \mathcal M_{j-1} + (1-\rho)M_jM_j^T \,\,\forall j\geq 1,
\end{split}
\label{eqn_the_KFAC_process}
\end{equation}
and can alternatively be written as $\mathcal M_k = \sum_{i=0}^k\kappa(i)\rho^{k-i}M_iM_i^T$. Ignoring the projection error of \textsc{rsvd} (it is very small for our purpose \cite{randomized_KFACS}), when performing \textsc{r-kfac} (with target rank $r$) instead of \textsc{k-fac} we effectively estimate $\mathcal M_k$ as
\begin{equation}
\begin{split}
&\tilde {\mathcal M}_{R;k,r} = U_{\mathcal M_k,r}  U_{\mathcal M_k,r}^T\mathcal M_kU_{\mathcal M_k,r}U_{\mathcal M_k,r}^T\,\,\,\forall k\geq 0,\,\,\,\text{where}
\\
U_{\mathcal M_k} D_{\mathcal M_k}  U_{\mathcal M_k}^T \,\,&\SVDeq\,\,\mathcal M_k =  \sum_{i=0}^k\kappa(i)\rho^{k-i}M_iM_i^T, \,\,\,\text{and}\,\,\, U_{\mathcal M_k,r} := U_{\mathcal M_k}[:,:r].
\end{split}
\label{eqn_the_R_process}
\end{equation}
Conversely, \textsc{b-kfac} effectively estimates $\mathcal M_k$ as $\tilde {\mathcal M}_{B,k}$, where $\tilde {\mathcal M}_{B,k}$ is given by
\begin{equation}
\begin{split}
 \tilde {\mathcal M}_{B,i+1}:=\rho  U_{\tilde {\mathcal M}_{B,i},r}&U_{\tilde {\mathcal M}_{B,i},r}^T\tilde {\mathcal M}_{B,i}U_{\tilde {\mathcal M}_{B,i},r}U_{\tilde {\mathcal M}_{B,i},r}^T+(1-\rho)M_{i+1}M_{i+1}^T\,\forall i\geq 0,
\\
\text{with}\,\tilde {\mathcal M}_{B,0} = M_0M_0^T,\,\,\,&U_{\tilde {\mathcal M}_{B,i}}D_{\tilde {\mathcal M}_{B,i}}U_{\tilde {\mathcal M}_{B,i}}^T\SVDeq\tilde {\mathcal M}_{B,i},\,\, U_{\tilde {\mathcal M}_{B,i},r}:=U_{\tilde {\mathcal M}_{B,i}}[:,:r];
\\
\text{we also define }& \mathcal B_i := U_{\tilde {\mathcal M}_{B,i},r}U_{\tilde {\mathcal M}_{B,i},r}^T\tilde {\mathcal M}_{B,i}U_{\tilde {\mathcal M}_{B,i},r}U_{\tilde {\mathcal M}_{B,i},r}^T\,\,\,\forall i\geq 0.
\end{split}
\label{eqn_the_B_process}
\end{equation}
Using equations (\ref{eqn_the_R_process})-(\ref{eqn_the_B_process}) one can easily compare the error (in K-factors) for \textsc{b-kfac} and \textsc{r-kfac}. The result is shown in \textit{Proposition 3.1}.
\begin{Proposition_3_1}
	For the quantities defined in equations (\ref{eqn_the_KFAC_process})-(\ref{eqn_the_B_process}) we have $\forall k$ that 
	\begin{equation}
	\norm{\mathcal M_k - \mathcal B_k}\geq\norm{\mathcal M_k - \tilde{\mathcal M}_{R,k,r}} \,\,\text{and}\,\,\,\norm{\mathcal M_k - \tilde{\mathcal M}_{B,k}}\geq\norm{\mathcal M_k - \tilde{\mathcal M}_{R,k,r+n_{\text{BS}}}},
	\end{equation}
	In any unitary-invariant norm.
\end{Proposition_3_1}
\textit{Proof.} \textit{Part 1:} Both $\mathcal B_k$ and $\mathcal M_{R,k,r}$ are rank $r$ matrices. By the properties of SVD, $\tilde {\mathcal M}_{R;k,r}$ is the optimal rank-$r$ truncation of $\mathcal M_k$ (that is, it has minimal error in any unitary-invariant norm; see \cite{RSVD_theory_error}, \cite{Optimality_of_SVD}). \textit{Part 2:} $\tilde{\mathcal M}_{B,k}$ is at most rank $r+n_{\text{BS}}$, and $\mathcal M_{R,k,r+n_{\text{BS}}}$ is rank $r+n_{\text{BS}}$. Apply similar reasoning to before. $\qed$

The interpretation of \textit{Proposition 3.1} is as follows. Both processes $\{\mathcal B_k\}_{k\geq0}$ and $\{\tilde {\mathcal M}_{R;k,r}\}_{k\geq0}$ construct a rank-$r$ estimate of $\mathcal M_k$. While $\{\tilde {\mathcal M}_{R;k,r}\}_{k\geq0}$ does this in an error-optimal way (w.r.t.\ unitary invariant norms), $\{\mathcal B_k\}_{k\geq0}$ will generally be suboptimal since $\mathcal B_k\ne \tilde{\mathcal M}_{R;k,r}$ will generally hold. Similarly, $\tilde{\mathcal M}_{B,k}$ and $\tilde {\mathcal M}_{R,k,r+n_{\text{BS}}}$ are rank $r+ n_{\text{BS}}$ estimates of $\tilde M_k$. Analogous reasoning follows.

\textit{Proposition 3.1} tells us two important things. Firstly, we see that the error of a \textsc{b-kfac} algorithm using a truncation rank of $r$, a batch-size of $n_{\text{BS}}$, and inverting based on $\tilde {\mathcal M}_{B,k}$ is lower bounded by the error of an \textsc{r-kfac} with target rank $r+n_{\text{BS}}$ and the same batch-size. Secondly, \textit{Proposition 3.1} tells us the best possible $\mathcal B_k$ is $\mathcal B_k= \tilde {\mathcal M}_{R,k,r}$. This raises scope for periodically ``refreshing'' $\mathcal B_k$ by setting $\mathcal B_k= \tilde {\mathcal M}_{R,k,r}$ through performing an \textsc{rsvd} of $\mathcal M_k$. We discuss this next.




\subsection{Brand RSVD K-FAC (B-R-KFAC)}
The discussion above raises a legitimate question: if within a \textsc{b-kfac} algorithm we perform an \textsc{rsvd} at some iteration $i>0$ and ``overwrite'' $\mathcal B_i = \tilde {\mathcal M}_{R,i,r}$, will this result in the errors $\forall k\geq i$ to be smaller than if we had not over-written $\mathcal B_i$?

\textit{Proposition 3.2} gives some intuition suggesting ocasionally overwriting $\mathcal B_i = \tilde {\mathcal M}_{R,i,r}$ in a \textsc{b-kfac} algorithm might be a good idea.
\begin{Proposition_3_2}
	For $j\geq 1$, let $\tilde {\mathcal M}^{R@i}_{i+j}$ and $\mathcal B^{R@i}_{i+j}$ be the $\tilde {\mathcal M}_{i+j}$ and $\mathcal B_{i+j}$ produced by process (\ref{eqn_the_B_process}) after over-writing $\mathcal B_i = \tilde {\mathcal M}_{R,i,r}$ at $i>0$. The error when doing so ($\forall q\geq 1$) is
	\begin{equation}
	E^{\text{R@i}}_{i+q}:= (\mathcal M_{i+q} - \tilde {\mathcal M}^{R@i}_{B,i+q}) = \rho^{q}(\mathcal M_i - \tilde {\mathcal M}_{R,i,r})+\sum_{j=1}^{q-1}\rho^{q-j}(\tilde{\mathcal M}^{R@i}_{B,i+j}-\mathcal B^{R@i}_{i+j}).
	\label{B_overwritten_err}
	\end{equation}
	When performing pure \textsc{b-kfac} we have the error at each $i+q$  ($\forall q\geq 1$) as
	\begin{equation}
	E^\text{(pure-B)}_{i+q}:= (\mathcal M_{i+q} - \tilde {\mathcal M}_{B,i+q}) = \rho^{q}(\mathcal M_i - \mathcal B_{i})+\sum_{j=1}^{q-1}\rho^{q-j}(\tilde{\mathcal M}_{B,i+j}-\mathcal B_{i+j}).
	\label{B_pure_err}
	\end{equation}
	Further, all the quantities within $(\cdot)$ are sym.\ p.s.d.\ matrices for any index $\geq 0$.
	
\end{Proposition_3_2}
\textit{Proof.} See appendix. $\qed$
 
\textit{Proposition 3.2} tells us that setting $l=1$ gives $E^\text{(pure-B)}_{i+1} = \rho(\mathcal M_i - \mathcal B_{i})$ and $E^{\text{R@i}}_{i+1}=\rho (\mathcal M_i - \tilde {\mathcal M}_{R,i,r})$, which combined with \textit{Proposition 3.1} gives $\norm{E^\text{(pure-B)}_{i+1}}\geq\norm{E^{\text{R@i}}_{i+1}}$. This tells us that performing the over-writing $\mathcal B_i = \tilde {\mathcal M}_{R,i,r}$ is certainly better for iteration $i+1$. But is it better for subsequent iterations?

Note that $\tilde{\mathcal M}^{R@i}_{B,i+j}-\mathcal B^{R@i}_{i+j}$ and $\tilde{\mathcal M}_{B,i+j}-\mathcal B_{i+j}$ are truncation errors at iteration $i+j$ along the ``overwritten'' and ``pure'' B-processes respectively. $\rho(\mathcal M_i - \tilde {\mathcal M}_{R,i,r})$ and $\rho(\mathcal M_i - \mathcal B_{i})$ are the \textit{initial} errors of these processes, when we set our starting-point at iteration $i+1$. Further note that since all the involved errors are symmetric p.s.d.\ matrices, all the terms in the sum have a positive contribution towards the norm of the total error (i.e.\ errors cannot ``cancel each-other out''). 

Equations (\ref{B_overwritten_err}) and (\ref{B_pure_err}) show that the contribution of the initial error towards $E^{(\cdot)}_{i+q}$ decays with $\uparrow q$ in both cases. Generally, one may construct examples where either one of $E^\text{(pure-B)}_{i+q}$ and $E^\text{R@i}_{i+q}$ have higher norms for $q\geq 2$. So we do not know how our ``overwritten'' process compares to the ``pure'' one for $q\geq 2$ (although one may argue the two converge as $q\to\infty$). Nevertheless, we can always overwrite $\mathcal B_{i+j}$ once again, and be sure this will give us at least another iteration on which our now twice overwritten process has better error than the ``pure'' one. This suggests that periodically overwriting $\mathcal B_i = \tilde {\mathcal M}_{R,i,r}$ by performing an \textsc{rsvd} every $T_{\text{RSVD}}$ steps \textit{may} lower the average \textsc{b-kfac} error. This is \textsc{b-r-kfac} (\textit{Algorithm 5}). Note that \textsc{b-r-kfac} mostly performs B-updates, so is cheaper than \textsc{r-kfac}.

\begin{algorithm}[H]
	\caption{Brand \textsc{rsvd} \textsc{k-fac} (\textsc{b-r-kfac})}
	Replace \textit{lines 12 - 13 in Algorithm 1} with:

		\If(\tcp*[h]{Time to over-write ``$\mathcal B_{k-1}$''}){$k \, \%\,T_{\text{RSVD}}==0$}
		{
			$\tilde U^{(l)}_{A,k-1}\tilde D^{(l)}_{A,k-1} (\tilde U^{(l)}_{A,k-1})^T = \textsc{rsvd}(\bar {\mathcal A}^{(l)}_{k-1})$
			
			$\tilde U_{\Gamma,k-1}^{(l)}\tilde D_{\Gamma,k-1}^{(l)} (\tilde U^{(l)}_{\Gamma,k-1})^T = \textsc{rsvd}(\bar \Gamma^{(l)}_{k-1})$
		}
		\Else(\tcp*[h]{Use standard \textsc{b-kfac} truncation to get ``$\mathcal B_k$''}){
			Truncate as in \textit{lines 2 - 4} of \textit{Algorithm 4}
		}
		
		Do \textit{lines 5 - 7} of \textit{Algorithm 4} \tcp*{(Perform B-update)}
	
\end{algorithm}

\subsubsection{Why use $\tilde {\mathcal M}_{B,k}$, not $\mathcal B_k$?} Consider $\mathcal M_k - \mathcal B_k = (\mathcal M_k - \tilde{\mathcal M}_{B,k}) + (\tilde{\mathcal M}_{B,k}- \mathcal B_k)$.
\textit{Proposition 3.2} tells us both $ (\mathcal M_k - \tilde{\mathcal M}_{B,k})$ and $ (\tilde{\mathcal M}_{B,k}- \mathcal B_k)$ are sym-p.s.d. matrices. Thus, adding $(\tilde{\mathcal M}_{B,k}- \mathcal B_k)$ to $ (\mathcal M_k - \tilde{\mathcal M}_{B,k})$ cannot decrease the latter's singular values. Therefore, we have $\norm{\mathcal M_k - \mathcal B_k} \geq \norm{\mathcal M_k - \tilde{\mathcal M}_{B,k}}$ in any norm that can be expresed purely in terms of singular values.


\subsection{Lighter Correction of B-KFAC, and the B-KFAC-C algorithm}
Periodically overwriting $\mathcal B_i = \tilde {\mathcal M}_{R,i,r}$ may impove \textsc{b-kfac}, but the over-writing operation is expensive, since it employs an \textsc{rsvd} (of target rank $r$). A cheaper alternative, is to perform a \textit{correction} where we improve the accuracy in \textit{only} $n_{crc} < r$ modes of our current \textsc{b-kfac} representation as shown in \textit{Algorithm 6}. 

\begin{algorithm}[H]
	\caption{Light Correction to \textsc{b-kfac} representations}
	\textbf{Input:} $\tilde U^{(l)}_{\cdot,k},\,\tilde D^{(l)}_{\cdot,k}$ (the \textsc{b-kfac} representation), EA K-Factor ${\mathcal M}_k$, $n_{crc}$
	
	col\_idx $= \text{random\_choice}(r,n_{crc})$  \tcp{choose $n_{crc}$ rows out of the $r$ without replcement}
	
	\tcp{Now project ${\mathcal M}_k$ on chosen random subspace}
	${\mathcal M}_{S,k} = \big(\tilde U^{(l)}_{\cdot,k}[:,\text{col\_idx}]\big)^T {\mathcal M}_k\big(\tilde U^{(l)}_{\cdot,k}[:,\text{col\_idx}]\big)$\tcp{$\in\mathbb R^{n_{crc}\times n_{crc}}$}
	
	$U D U^T =\text{symmetric\_EVD}({\mathcal M}_{S,k})$ 
	
	\tcp{Correct $\tilde U^{(l)}_{\cdot,k},\,\tilde D^{(l)}_{\cdot,k}$ in the subspace described by col\_idx}
	$\tilde U^{(l)}_{\cdot,k}[:,\text{col\_idx}] = U$; 	$\tilde D^{(l)}_{\cdot,k}[:,\text{col\_idx}] = D$

	\textbf{Outut:} $\tilde U^{(l)}_{\cdot,k},\,\tilde D^{(l)}_{\cdot,k}$ (a more accurate \textsc{b-rsvd} representation)

\end{algorithm}

The correction enforces that the projection of our new \textsc{b-rsvd} representation (in \textit{line 8}) on our randomly chosen  $n_{crc}$-dimensional subspace of $\tilde U_{\cdot,k}^{(l)}$ (in \textit{line 2}) be the same as the one of the true EA K-factor ${\mathcal M}_k$. Performing a correction at $k$ can only reduce the error $\norm{\mathcal M_k - \tilde {\mathcal M}_k}_F$, but not\footnote{Consider $E_k = \mathcal M_k - \tilde {\mathcal M}_k$. For any matrix we have: $\norm{E_k}^2_F = \norm{U^TE_kU}^2_F + \norm{U_\perp^TE_kU}^2_F + \norm{U^TE_kU_\perp}^2_F + \norm{U_\perp^TE_kU_\perp}^2_F$ when the matrix $[U, U_{\perp}]\in\mathbb R^{d\times d}$ is orthogonal (thus $U$ orthonormal). Performing the correction ensures $U^TE_kU = U^T(\mathcal M_k - \tilde {\mathcal M}_k)U=0$ - but for our pre-correction error $\norm{U^TE_kU}^2_F\geq 0$.} increase it. Similarly to the \textsc{rsvd}-based overwriting, it is unclear whether the effect on future iterations is surely positive. Note that we apply the correction to $\mathcal M_k$ and not to $\mathcal B_{k-1}$. Similarly to over-writing, we apply the correction with a smaller frequency than the one of the \textsc{b-kfac} update.

By using the lighter correction instead of the more expensive over-writing of $\mathcal B_k$, we can reduce our computational cost from $\mathcal O(d^2(r+r_o)^2)$ to $\mathcal O(d^2 n_{crc} + n_{crc}^4)$. This is substantially better if we choose $n_{crc}\leq0.5 r$. 
\startsquarepar We prefer selecting columns of $\tilde U^{(l)}_{\cdot,k}$ at random rather than picking its largest modes for 2 reasons. First, after multiple consecutive Brand updates (and no correction / overwriting) it is possible that the largest singular modes of $\mathcal M_k$ are along directions of relatively low singular values of the \textsc{b-kfac} representation. \stopsquarepar

Second, always picking the largest singular modes of the \textsc{b-kfac} representation would tend to give us scenarios where we always correct the same modes. This comes from the fact that both the EA K-factor and the incoming update are positive semi-definite, and thus the \textsc{b-kfac} representation can only underestimate singular-values, but not overestimate them.

\begin{algorithm}[H]
	\caption{Corrected Brand \textsc{k-fac} (\textsc{b-kfac-c})}
	Insert the following 2 lines after \textit{line 7} of \textit{Algorithm 4}:
	
	\If(\tcp*[h]{Time to correct ``$\mathcal M_{k}$''}){$k \, \%\,T_{\text{corct}}==0$}
	{
		Perform \textit{Algorithm 6} to $\{(\tilde U^{(l)}_{A,k}\tilde D^{(l)}_{A,k})\}_l$ and $\{(\tilde U_{\Gamma,k}^{(l)},\tilde D_{\Gamma,k}^{(l)})\}_l$
	}
	
\end{algorithm}
\noindent \textbf{Hyperparameters Note:} In practice we use the parameter $\phi_{crc}:=n_{crc}/r$.

\subsection{A mixture of Randomized KFACs and Brand New KFACs}
Recall our discussion in \textit{Section 2.1} about K-Factors dimensions. In practice, we have $n^{(l)}_M > d^{(l)}_{\mathcal A}$ and $n^{(l)}_M > d^{(l)}_{\Gamma}$ for convolutional layers, but $n^{(l)}_M = n_{\text{BS}}<\min(d^{(l)}_{\mathcal A}, d^{(l)}_{\Gamma})$ (or even $n^{(l)}_M = n_{\text{BS}}\ll\min(d^{(l)}_{\mathcal A}, d^{(l)}_{\Gamma})$) for FC layers. This means \textsc{b-kfac} will only save computation time (relative to \textsc{r-kfac} or \textsc{sre-kfac}) for the FC layers, and will be slower for Conv layers\footnote{While in practice we might still have $n^{(l)}_M < d_{\mathcal A}^{(l)}$ and/or $n^{(l)}_M < d_{\Gamma}^{(l)}$ (in whihc case we could still apply the B-update), in this paper we assume that's not the cae, for simplicity.}. This issue is simply solved by using \textsc{b-kfac}, \textsc{b-r-kfac}, or \textsc{b-kfac-c} for the FC layers only, and \textsc{r-kfac} or \textsc{sre-kfac} (\cite{randomized_KFACS}) for the Conv layers. When the FC layers are very wide, becoming the computational bottle-neck, speeding up the FC layers computation can give substantial improvement. 


\subsubsection{Spectrum Continuation} 
Both randomized \textsc{k-fac} algorithms (eg.\ \textsc{r-kfac}) and the \textsc{b-kfac} variants we propose here are effectively setting $d-r$ (where $r$ is the rank of the K-Factor estimate) eigenvalues to zero\footnote{We talk about the matrices we have before regularization with ``$+\lambda I$''.}. In reality, we know that the EA K-Factors eigen-spectrum typically decays gradually, rather than have an abrupt jump (\cite{randomized_KFACS}), and we also know all eigen-values are non-negative. Using this information one may try to correct the missing eigen-tails. 

A quick fix is to say all the missing eigenvalues are equal to the minimal one available. Using this trick, we observed slightly better performance for all algorithms (\textsc{r-kfac} and all \textsc{b-kfac} variants). This is probably because over-estimating the eigenspectrum is better than underestimating it, since it gives more conservative steps. This spectrum continuation trick is implementable by replacing $\lambda\leftarrow\lambda + \min_i D_k[i,i]$ and  $D_k\leftarrow D_k - (\min_i D_k[i,i]) I$ in \textit{lines 16-17} of \textit{Algorithm 1}. The replacements also affect all proposed algorithms, as these merely amend \textit{lines 12-13} of \textit{Algorithm 1}. We use this trick for all layers.

\subsubsection{B-KFAC is a low-memory K-FAC} \textsc{b-kfac} never needs to form any (large, square) K-factor, and only ever stores skinny-tall matrices (large height, small width). Thus, \textsc{b-kfac} can be used as a \textit{low-memory version} of \textsc{k-fac} or \textsc{r-kfac} when these would overflow the memory due to forming the K-Factor. We cannot use \textsc{b-r-kfac} and \textsc{b-kfac-c} as low-memory, as they require K-Factor formation.


\section{Error Analysis: Approximate K-Factor Inverse Updates}
\subsection{Theoretical Comparison of R-KFAC and B-R-KFAC errors}

Based on \textit{Proposition 3.2} we argued that, given a \textsc{b-kfac} algorithm, one might expect that periodically over-writing $\mathcal B_k$ with the rank-$r$ \textsc{r-kfac} estimate $\tilde {\mathcal M}_{R,k,r}$ (by performing an \textsc{rsvd} on $\mathcal M_k$) might give better error $\norm{\mathcal M_j - \tilde {\mathcal M}^{R@T_{\text{RSVD}}i}_{B,j}}$ for all iterations (but it was not guaranteed). This previous comparison was between \textsc{b-kfac} and \textsc{b-r-kfac}, and it represented our motivation behind \textsc{b-r-kfac}.

In this segment, change our point of view and think about what happens if we take a given \textsc{r-kfac} algorithm with $[T_{\text{inv}}/T_{\text{updt}}]=:R_\nu\in\mathbb Z^+\setminus\{1\}$, and introduce B-updates (to the inverse estimates) each time the K-Factors are updated, but the RSVD inverse is not recomputed.  Since we are only interested in the K-factors and not the optimization steps, we can take $T_{\text{updt}}=1$  w.l.o.g., so $T_{\text{inv}}=R_\nu$. This point of view amounts to comparing \textsc{r-kfac} ($T_{\text{inv}}=R_\nu$) with \textsc{b-r-kfac} ($T_{RSVD}= R_\nu$), where new K-factor information comes every iteration. Thus, we have to compare the error of \textit{performing no update} versus the error of \textit{performing \textsc{b}-updates}, starting from an RSVD update at $k=0$. \textit{Proposition 4.1} tells us what the error\footnote{Measured as the difference between the true EA-Kfactor and the approximate one used to cheaply compute the inverse.} is for \textit{\textsc{b}-update}, as well as for \textit{no update}.
\begin{Proposition_4_1}
	Let $\tilde {\mathcal M}_k$ be an approximation of $\mathcal M_k$ which is obtained by performing an \textsc{r-kfac} update at $k=0$, and either no other update thereafter, or \textsc{b}-updates (every step) thereafter. The error in $\mathcal M_k$ when using one of these approximations is of the form
	\begin{equation}
	\mathcal M_k - \tilde {\mathcal M}_{k} = \sum_{i=0}^{k} \kappa(i)\rho^{k-i}E_i,\,\,\,\,\text{with}\,\,\,E_0 =(\mathcal M_0 - \tilde {\mathcal M}_{R;0,r}).
	\label{prop5_1_general_err_form}
	\end{equation}
	When performing \textsc{rsvd} initially (at $k=0$), and no update thereafter we have 
	\begin{equation}
	E_j =  M_j M_j^T - \tilde {\mathcal M}_{R;0,r}.
	\label{err_with_no_update_performed}
	\end{equation}
	When performing \textsc{rsvd} initially (at $k=0$), and \textsc{b}-updates thereafter we have
	\begin{equation}
	E_i = \frac{1}{1-\rho}(\tilde{\mathcal M}_{j} - \mathcal B_i)\,\,\forall i\in\{1,...,k-1\},\,\,\,\text{and}\,\,\,E_k = 0,
	\label{err_with_B_update_performed}
	\end{equation}
	and $\tilde{\mathcal M_i} = \tilde {\mathcal M}_{B,i}$.  where $\mathcal B_i$ is as in (\ref{eqn_the_B_process}).
\end{Proposition_4_1}
\textit{Proof.} Trivial - see appendix. $\qed$
\newpage

\textbf{\textit{Importantly}}, Note that $\tilde{\mathcal M}_{i} - \mathcal B_i$ is the (s.p.s.d.) truncation error matrix when optimally truncating $\tilde{\mathcal M_i} = \tilde {\mathcal M}_{B,i}$ to rank $r$ (follows from (\ref{eqn_the_B_process})).

\textit{Proposition 4.1} shows that the over-all error is an exponential average of the errors $\{E_j\}_{j\geq 0}$. Note that $E_0$ is the same in both cases, but the errors arising for $j\geq 1$ are different in the two cases. Clearly, as more steps are taken (without any \textsc{rsvd} again), the overall error will depend less on the initial error $E_0$.

The error in (\ref{err_with_no_update_performed}) is revealing - it tells us that when no update is performed, we obtain the estimates $\tilde {\mathcal M}_k$ by pretending the incoming terms $M_iM_i^T$ are the same as our current EA K-factor estimate (i.e.\ by pretending $M_iM_i^T = \tilde {\mathcal M}_{R,0,r}$ $\forall j\geq 0$; note that $\tilde {\mathcal M}_{R,0,r}$ is the optimal rank-$r$ truncation of $\mathcal M_0= M_0M_0^T$). 

The error $E_i$ ($i\geq 1$) when \textsc{b}-updates are performed (in (\ref{err_with_B_update_performed})) is the (scaled) truncation error when optimally-truncating the matrix $\tilde {\mathcal M}_{B,i}$ (of maximal rank $r+n$) down to rank $r$. Importantly, $E_i$ does not depend on previous truncation errors, but only on the truncation error at iteration $i$. This simple error decomposition in the case of \textsc{b}-updates is essential for our following results.

In general\footnote{Altough not always.}, one might expect that the truncation error (i.e.\ for \textsc{b}-updates) is smaller than the error introduced by pretending $M_iM_i^T =  \tilde {\mathcal M}_{R,0,r}$ (i.e. by doing nothing for $i\geq 1$). While proving such probabilistic bounds is theory-heavy, we can easily show that there exists at least one case where $\norm{E_i}_F$ for no-update is larger than the upperbound of $\norm{E_i}_F$ ($i\geq 1$) for \textsc{b}-update. That is, the worst-case scenario when performing \textsc{b}-updates is surely better than the worst-case scenario when doing no updates. The results are summarised in \textit{Proposition 4.2}.
\begin{Proposition_4_2}
	When performing \textsc{rsvd} initially (at $k=0$), and no updates thereafter $\norm{E_j}_F$ can get as high as
	\begin{equation}
	\norm{E_j}_F =  \sqrt{\norm{M_jM_j^T}^2_F + \norm{\tilde {\mathcal M}_{R,0,r}}^2_F}, \,\,\,\forall j\in\{1,...,k\}.
	\label{eqn_1_to_prove_in_prop_4_2}
	\end{equation}
	When performing \textsc{rsvd} initially, and \textsc{b}-updates thereafter $\norm{E_j}_F$ is bounded as:
	\begin{equation}
	\norm{E_j}_F\leq \norm{M_jM_j^T}_F, \,\,\,\forall j\in\{1,...,k-1\}, \,\,\,\text{and}\,\,\, E_k = 0.
	\label{eqn_2_to_prove_in_prop_4_2}
	\end{equation}
\end{Proposition_4_2}
\textit{Proof.} See appendix. $\qed$

Note that the norm of $\tilde {\mathcal M}_{R,0,r}$ is always positive. Using (\ref{prop5_1_general_err_form}) and triangle inequality, we see that the overall EA K-factor error norm has the upper bound
\begin{equation}
\norm{\mathcal M_k - \tilde {\mathcal M}_k}_F \leq \rho^k\norm{\mathcal M_0 - \tilde {\mathcal M}_{R,0,r}}_F + (1-\rho)\sum_{j=1}^{k}\rho^{k-j}\norm{E_j}_F.
\end{equation}
Now, using \textit{Proposition 4.2}, we see that the overall error is better \textit{under the worst case scenario} for B-updates than under the worst-case scenario for no-updates.
\subsection{Numerical Error Investigation: Experimental Set-up}
We now look at the error of \textsc{b}-updates numerically as a way of complementing the theoretical results. To do so, we consider the following setup. We fix the frequency at which the updates to K-Factors are incoming (i.e.\ fix $T_{\text{updt}}$; here we set $T_{\text{updt}}=10$). For our fixed $T_{\text{updt}}$, a \textsc{k-fac} algorithm with $T_{\text{inv}} = T_{\text{updt}} = 10$ always maintains the inverse K-factors at their exact values. Thus, we take this to be the benchmark in our numerical error measurements. We then ask: what is the error between an algorithm which does not hold the exact value of inverse K-factors (for example an \textsc{r-kfac}, a \textsc{b-kfac}, or a even \textsc{k-fac} with $T_{\text{inv}}>T_{\text{updt}}$) and the benchmark? In principle, we could compute this error at each and every step. However, doing so is very expensive. We thus choose to only compute the error for two sequences\footnote{Since the eigenspectrum-decay in K-Factors is not significant until epoch 10-15 (see \cite{RSVD_paper}), the errors of both \textsc{r-kfac} and variants of \textsc{b-kfac} are relatively large initially - but should be (and were) relatively small and constant from epoch 15 onwards.} of 300 consecutive steps - one starting at epoch 15, and one starting at epoch 30. This is sufficient to draw conclusions.


Many \textbf{error metrics} could be used. We consider four which we believe are most relevant: \textit{(1) Norm Error in $\mathcal A^{-1}$}: $\norm{\tilde{\mathcal A}_k^{-1} - \mathcal A^{-1}_{k;\text{(ref)}}}_F\, /\, \norm{ \mathcal A^{-1}_{k;\text{(ref)}} }_F$, \textit{(2) Norm Error in $\Gamma^{-1}$:} $\norm{\tilde \Gamma_k^{-1} - \Gamma^{-1}_{k;\text{(ref)}}}_F\, /\, \norm{ \Gamma^{-1}_{k;\text{(ref)}} }_F$, \textit{(3) Norm Error in Subspace Step:} $\norm{\tilde{s}_k - s_{k;\text{(ref)}}}_F\, /\, \norm{s_{k;\text{(ref)}}}_F$, \textit{(4) Angle Error in Subspace Step:} $1 - \cos(\angle[\tilde{s}_k,{s}_{k;\text{(ref)}} ] )$. Here, quantities marked with \textit{tilde} represent the ones of our approximate algorithms, while quantities marked with \textit{``ref''} represent the ones of the reference (benchmark) algorithm (\textsc{k-fac} with $T_{\text{inv}} = T_{\text{updt}}=10$).

Recall that, our proposed algorithm focuses on FC layers. The network architecture\footnote{The learning problem is CIFAR10 classification with slightly ammended VGG16\_bn.} we use is the one given in \textit{Section 6}. This only has two FC layers, out of which it only makes sense to perform B-updates for the first FC layer. Thus, our error metrics will only relate to this first FC layer (marked as \textit{``FC layer 0''} in figures). Note that the steps $s_k$ considered in the above paragraph are the subspace steps (in the \textit{FC layer 0} parameters subspace), thus slightly overwriting the notation, just for this section.

The \textbf{algorithms we consider} are the ones introduced in \textit{Section 3}: (1) \textsc{b-kfac} with $T_{\text{Brand}} = 10$; (2) \textsc{b-r-kfac} with $T_{\text{Brand}} = 10$, $T_{\text{RSVD}} = 50$, (3) \textsc{b-kfac-c} with $T_{\text{Brand}} = 10$; $T_{corct} = 50$, $\phi_{corct} = 0.5$; (4) \textsc{r-kfac} with $T_{\text{inv}} = 50$; (5) \textsc{r-kfac} with $T_{\text{inv}} = 10$; (6) \textsc{r-kfac} with $T_{\text{inv}} = 300$; (7) \textsc{k-fac} with $T_{\text{inv}} = 50$. For all these algorithms, new K-factor data is incoming with period $T_{\text{updt}}=10$. All unspecified algorithms hyper-parameters are as described in \textit{Section 6}. Note that \textsc{r-kfac} with $T_{\text{inv}}=300$ is meant to show how the error would increase with $k$ if no update is performed to the inverse K-factors - which are initially estimated in the \textsc{r-kfac} style. Comparing \textsc{r-kfac} $T_{\text{inv}}=300$ and \textsc{b-kfac} directly relates to the theoretical result in \textit{Section 4.1}. Other comparisons also give insights.

Note that for all algorithms based on low-rank truncation, the eigen-spectrum is continued as explained in \textit{Section 4}. We always start our sequence of 300 steps (over which the error is measured) exactly when the heaviest update of the algorithm at hand is performed. For this reason, the error measuring of \textsc{r-kfac} with $T_{\text{inv}}=300$ starts slightly later in the epoch. However since the eigenspectrum profile varies very slowly with $k$, this difference is immaterial.

\begin{figure}[h]
	\centering

	\includegraphics[trim={0.2cm 0.49cm 1.25cm 0.5cm},clip,width=0.499\textwidth]{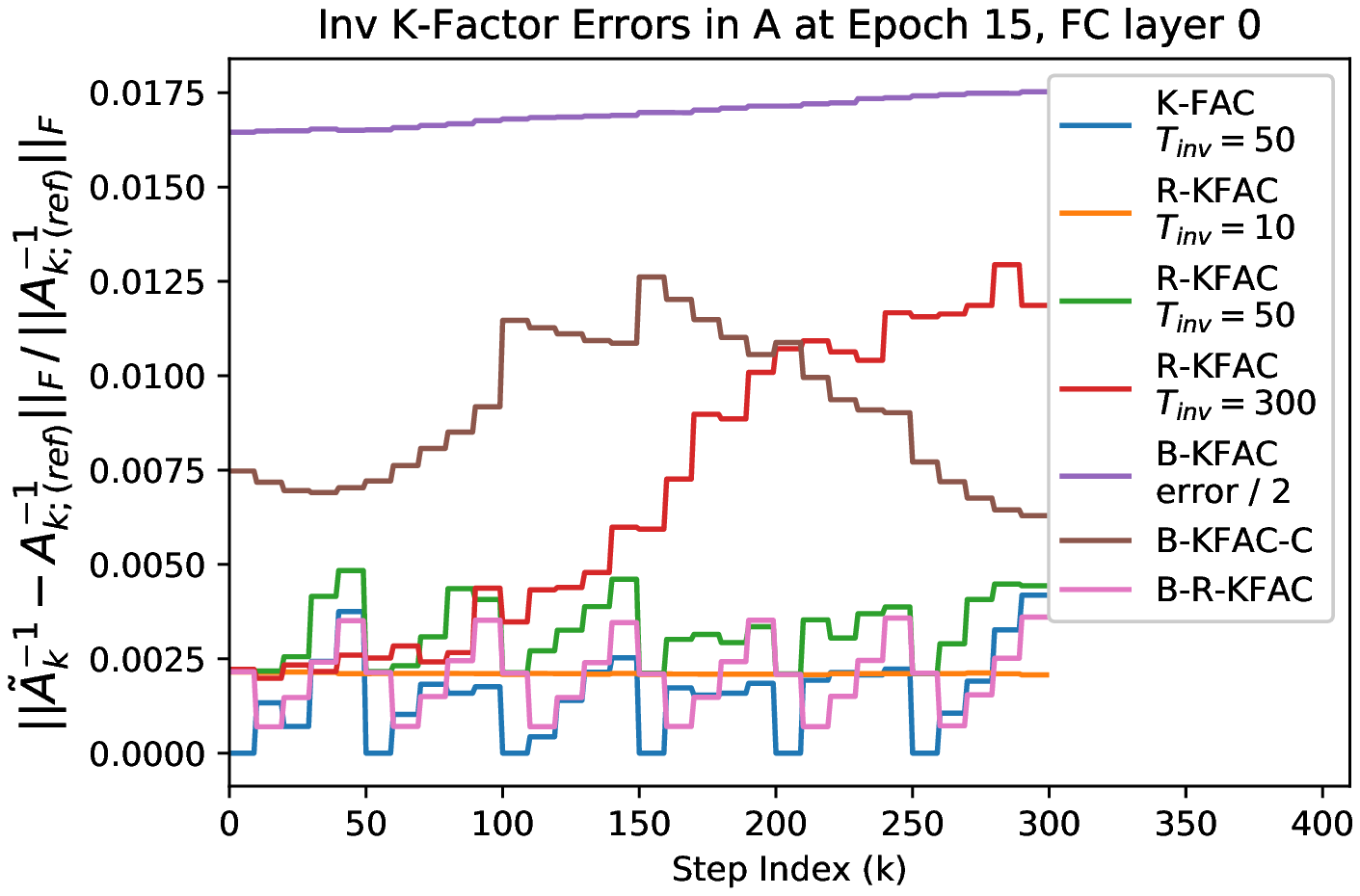}
	\includegraphics[trim={0.1cm 0.1cm 1.2cm 0.4cm},clip,width=0.493\textwidth]{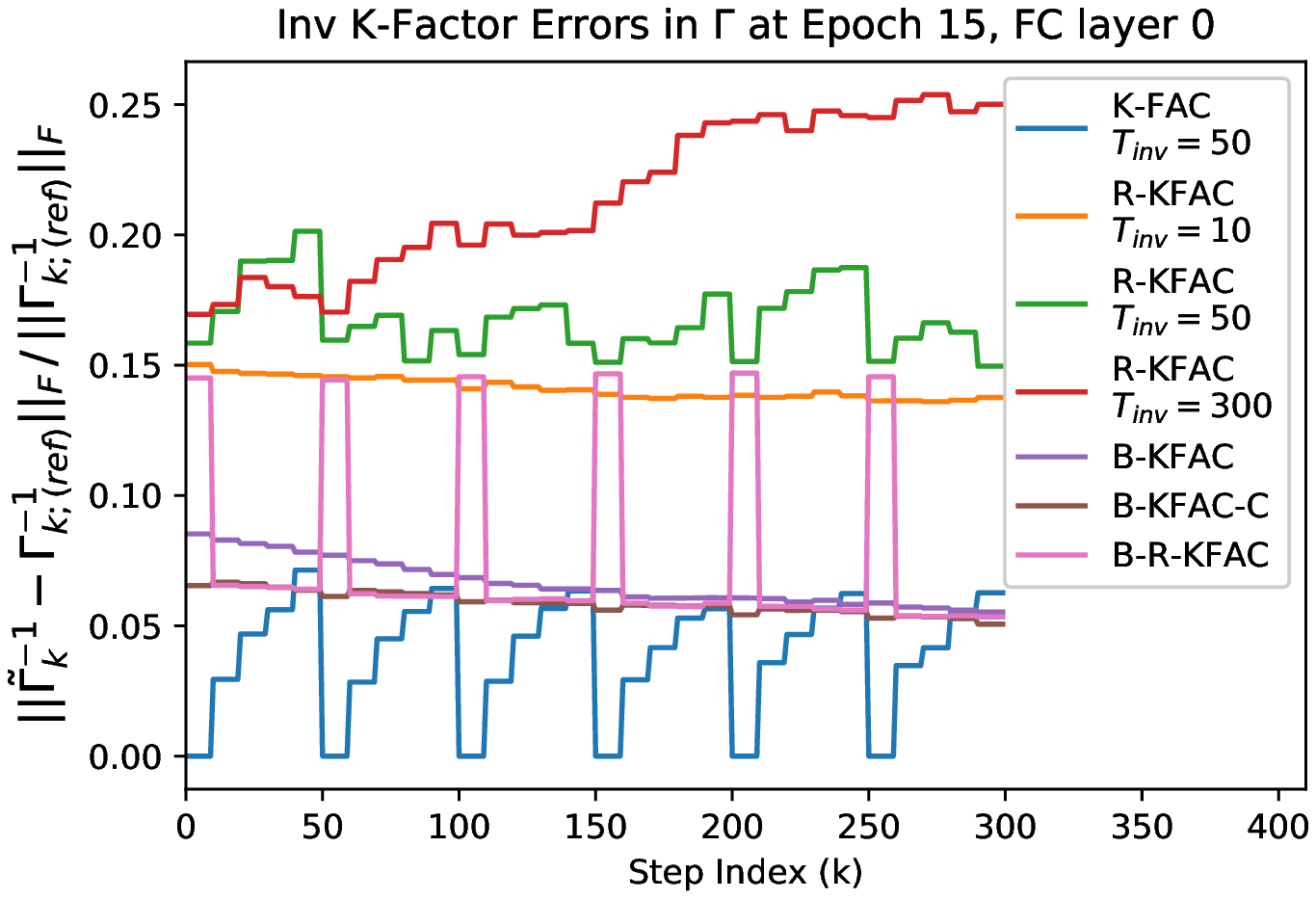}
	
	\includegraphics[trim={0.2cm 0.49cm 1.25cm 0.5cm},clip,width=0.499\textwidth]{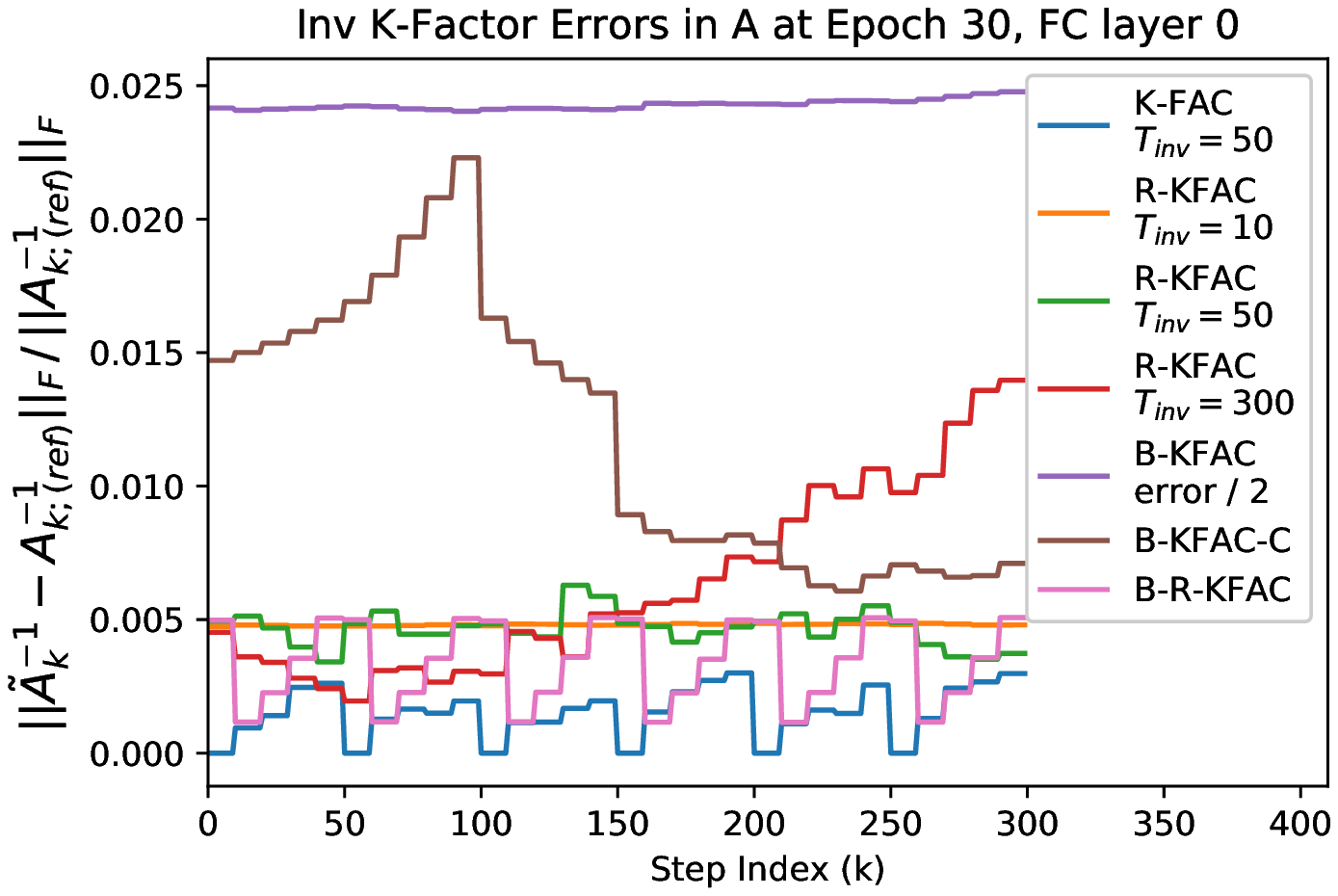}
	\includegraphics[trim={0.1cm 0.1cm 1.2cm 0.4cm},clip,width=0.493\textwidth]{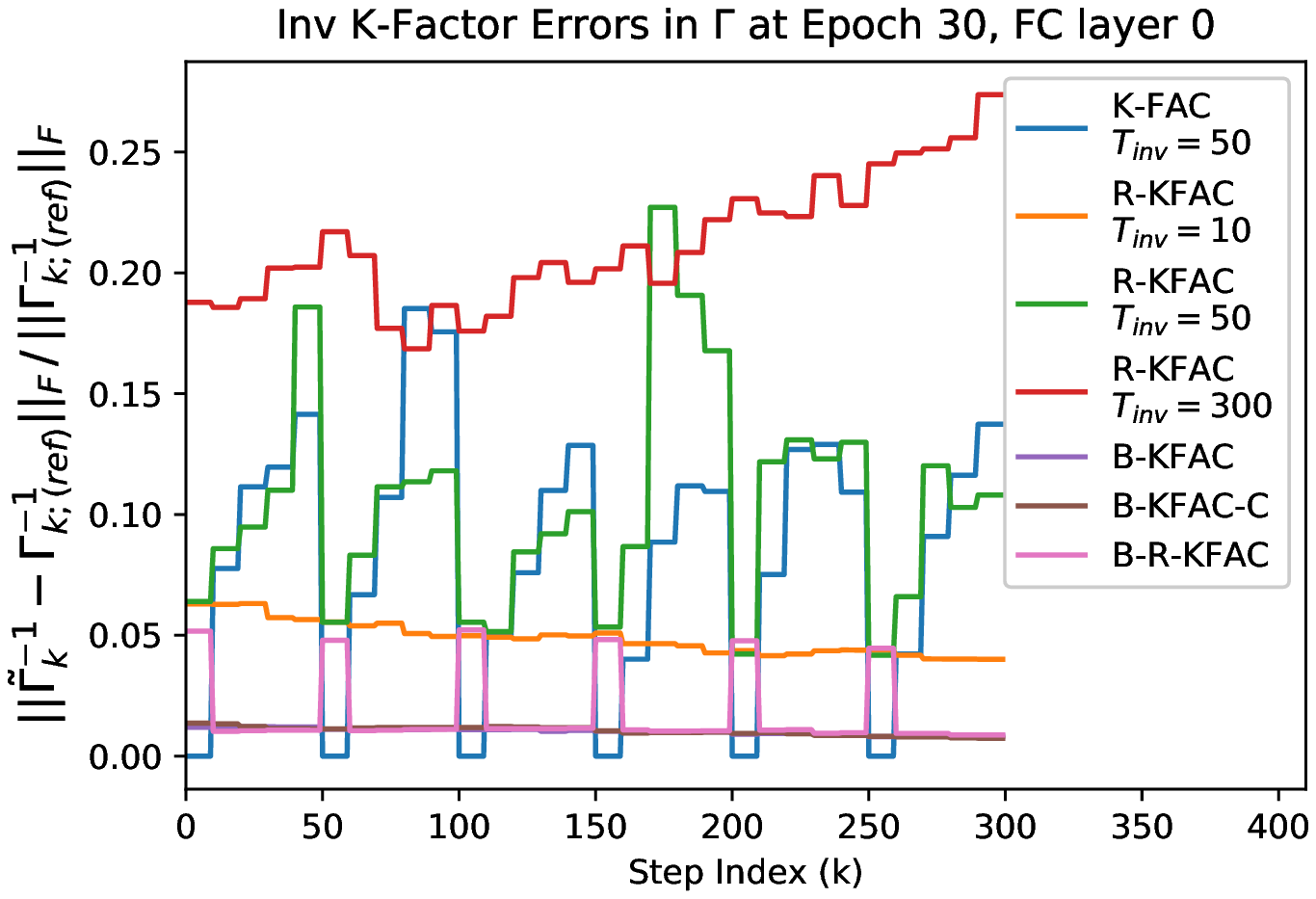}
	
	\caption{Error in inverse K-Factors. Each curve is the average over 5 runs.} 
	\label{err_in_inverse_Kfactors}
\end{figure}
\subsection{Numerical Error Investigation: Results}
\textit{Figure 1} shows the error metrics (1) and (2). \textit{Figure 2} shows the error metrics (3) and (4). Error metric and epoch vary across columns and rows respectively.
\vspace{+1ex}

\noindent \textbf{Error Periodicity and ``Reset Times''.} Note that aside from \textsc{b-kfac} and \textsc{r-kfac} with $T_{\text{inv}}=300$, all other algorithms have a period of 50 steps. This arises because these algorithms have a heavier update every $50$ step (either an \textsc{rsvd}-overwriting of ``$\mathcal B_k$'', or a correction), and a lighter (or no) update every 10 steps. The numerical results show that performing an over-writing  of ``$\mathcal B_k$'' (see \textsc{b-r-kfac}) or an EVD recomputation of inverse K-factor (see \textsc{k-fac} $T_{\text{inv}}=50$) always resets the error back to roughly the same level, while performing a correction reduces the error down to more variable levels (see \textsc{b-kfac-c}). This is intuitive: unlike the other two heavy updates, the correction's output depends on the approximate K-factor inverse representation to be corrected.

\vspace{+1ex}
\noindent \textbf{R-KFAC $T_{\text {inv}}=50$ vs K-FAC $T_{\text {inv}}=50$.} We see that the error patterns of these two algorithms are very similar. The error of \textsc{r-kfac} is only mildly larger than the one of \textsc{k-fac}, showing there is significant eigenspectrum decay (\cite{randomized_KFACS}).

\vspace{+1ex}
\noindent \textbf{B-R-KFAC vs R-KFAC: Relationship to \textit{Propositions 4.1} and \textit{4.2}.} There are two important comparisons to note here. First, comparing \textsc{b-r-kfac} and \textsc{r-kfac} with $T_{inv}=50$, we see that performing a \textsc{b-}update is almost always better than performing no update in terms of all error metrics and for all considered epochs (at least for the considered set-up). This result for the first two error metrics relates strongly\footnote{The relation is not perfect because the error metrics (1) and (2) consider the more practical error based on inverses, and uses spectrum continuation (see \textit{Section 3.5}).} to \textit{Proposition 4.2}, but further to the weaker result that the theory predicts, it shows that the error in K-Factors is almost always better when a \textsc{b-}update is performed than when no update is performed.

\begin{figure}[t]
	\centering

	\includegraphics[trim={0.0cm 0.1cm 1.5cm 0.4cm},clip,width=0.496\textwidth]{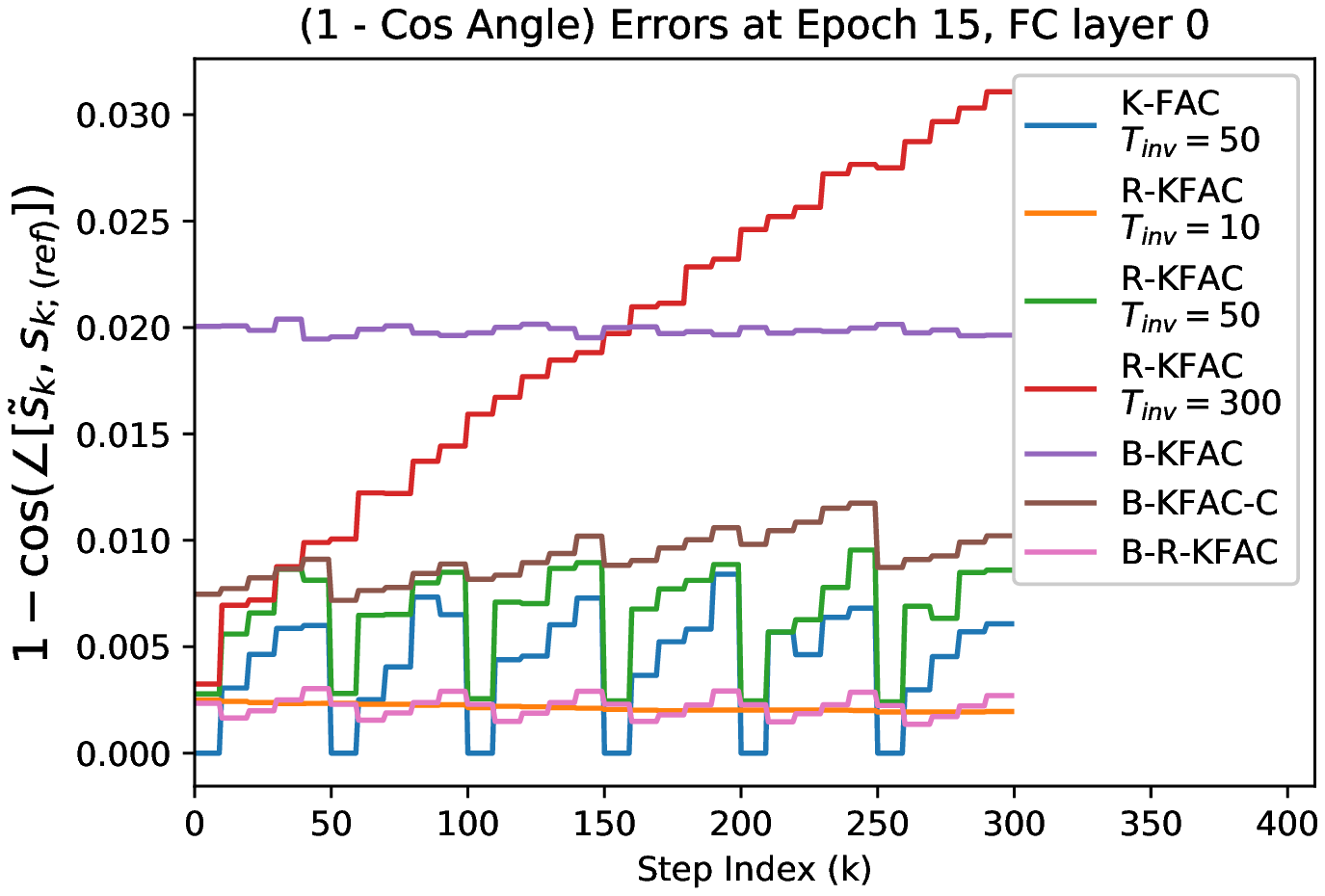}
	\includegraphics[trim={0.0cm 0.1cm 1.5cm 0.4cm},clip,width=0.496\textwidth]{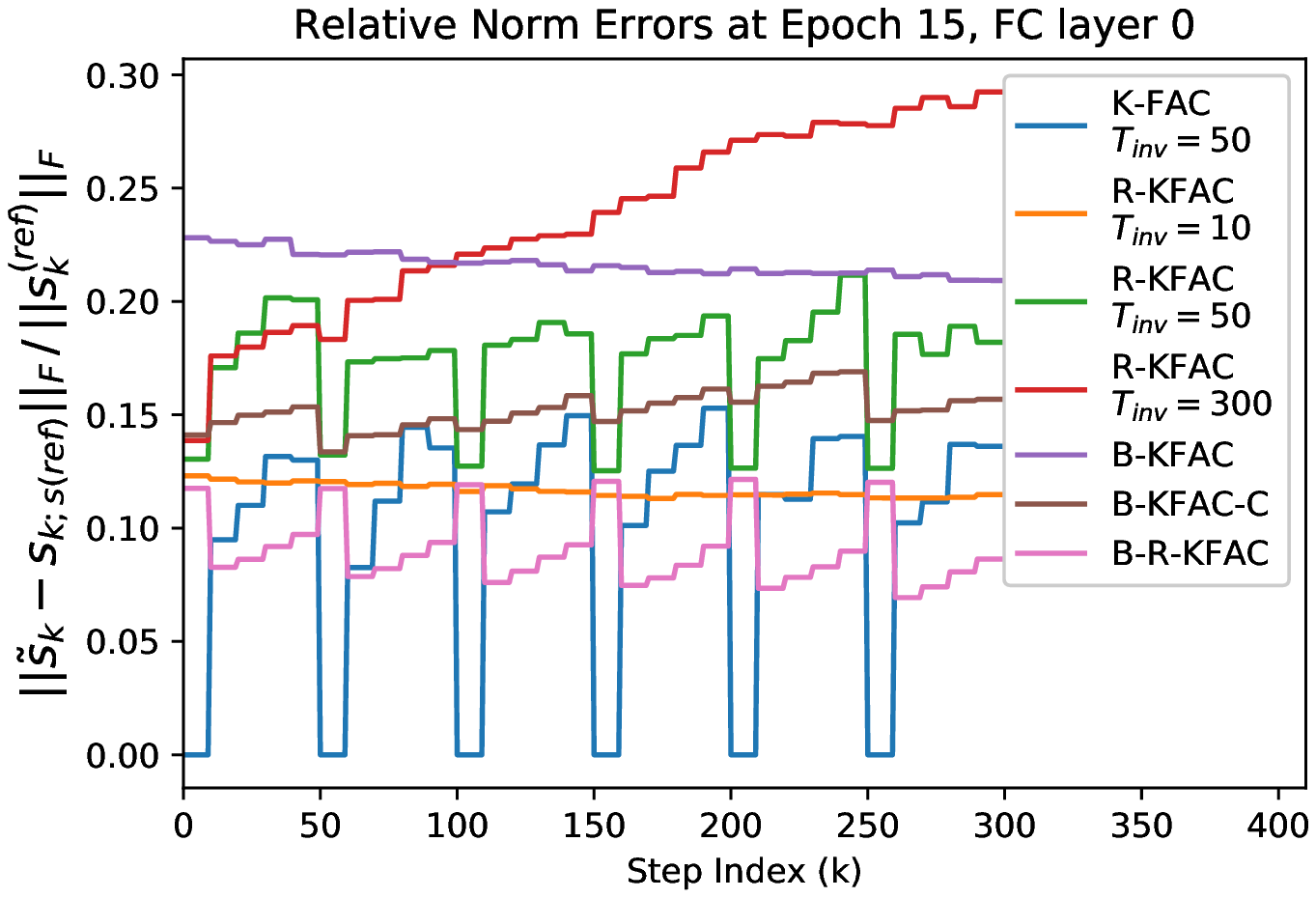}
	
	\includegraphics[trim={0.0cm 0.1cm 1.5cm 0.4cm},clip,width=0.496\textwidth]{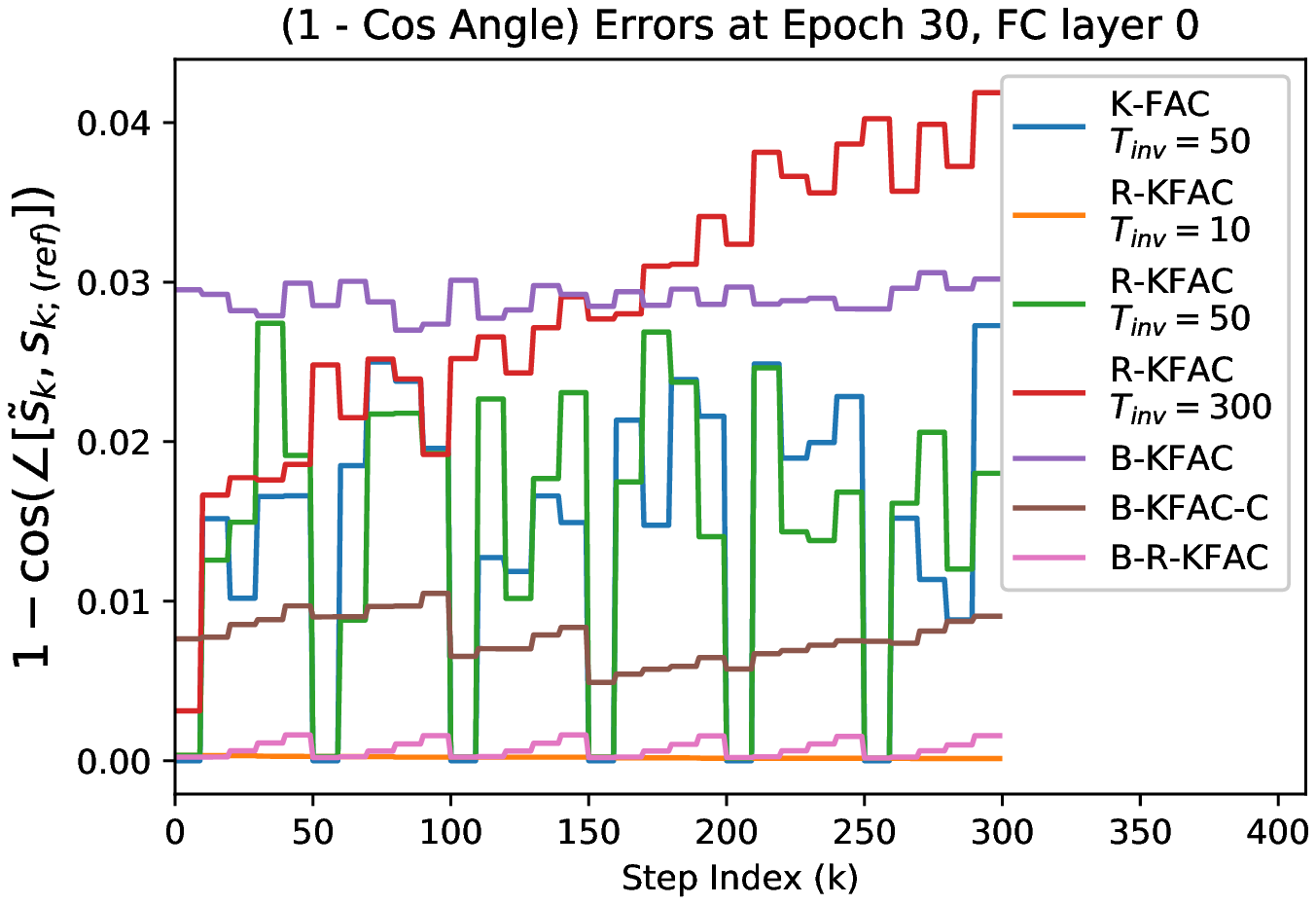}
	\includegraphics[trim={0.0cm 0.1cm 1.5cm 0.4cm},clip,width=0.496\textwidth]{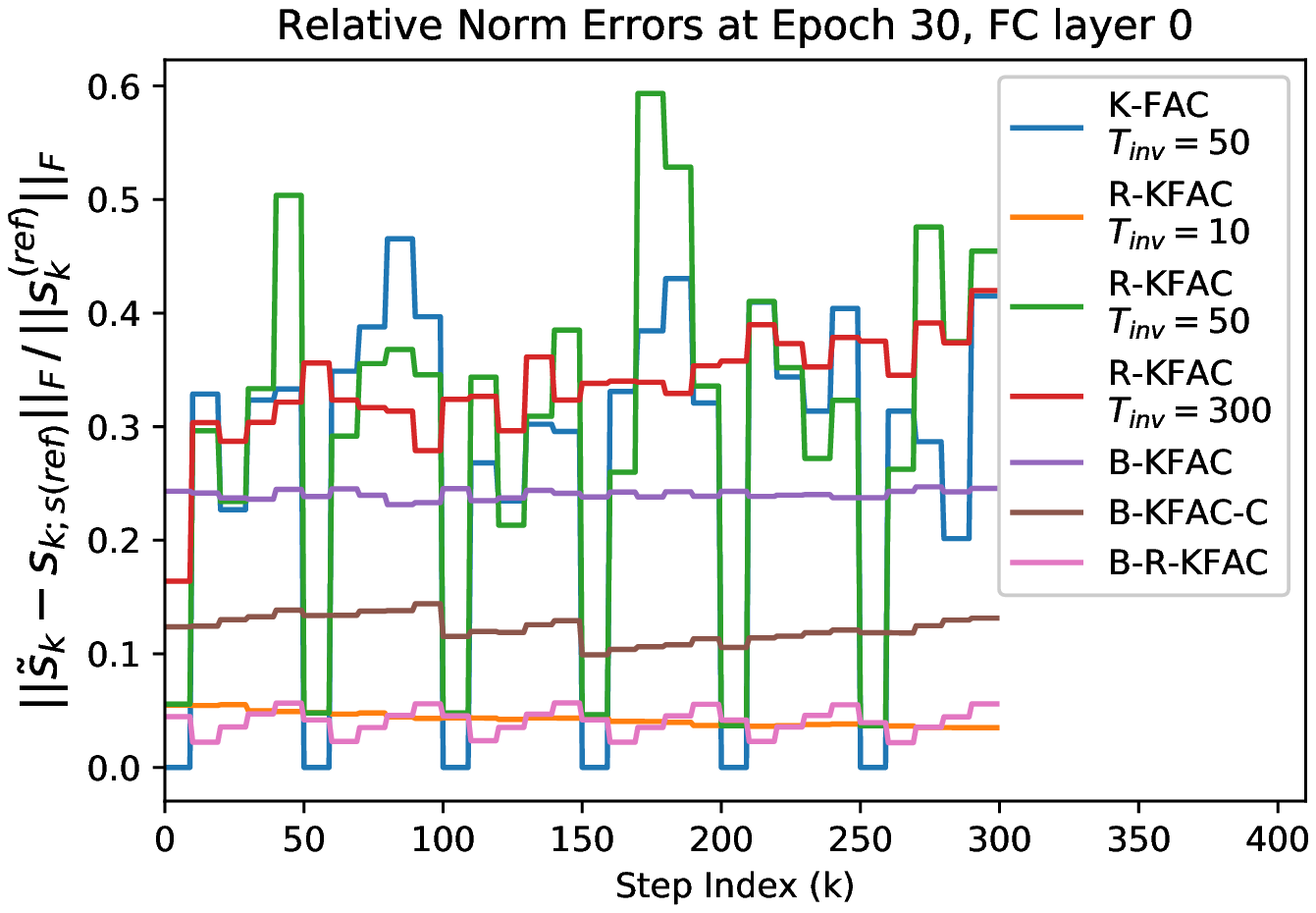}
	
	\caption{Error in Step. Since the gradient changes at each location, the error in step typically changes on every iteration, even though the error in K-Factor can change only once every 10 iterations. Thus, we effectively have 10 different samples from the distribution of step-error for each given $\{\text{approximate K-Factor representation, Benchmark K-Factor}\}$ pair. We report the average over these 10 samples only (giving us a constant error over iterations 200-209 for instance). As before, we further average over 5 runs.}
	\label{err_in_dir_and_angle}
	
	\vspace{-3ex}
\end{figure}

Secondly, comparing the results of \textsc{b-kfac} and \textsc{r-kfac} $T_{\text{inv}}=300$ is  effectively comparing the steady-state error of \textsc{b-kfac} with the error of "no reset" \textsc{r-kfac} (which performs one \textsc{r-kfac} update at the beginning of the examined period, and no updates thereafter). We see that for all error metrics, the error of  \textsc{b-kfac} is fairly constant while the error of \textsc{r-kfac} $T_{\text{inv}}=300$ grows fast. The error of the latter algorithm very rapidly exceeds the one of \textsc{b-kfac} for the second error metric. In relation to \textit{Proposition 4.1}, this suggests that $\norm{E_j}_F$ is much larger for no-update than for \textsc{b-}updates when $\mathcal M_k=\bar {\Gamma}_k$. While the same phenomenon did not occur on the 300 steps interval considered for error metric (1) and K-factor $\bar {\mathcal A}_k$, this would have occurred about 900 steps later.

From a more practical point of view, the step-related error metrics (in \textit{Figure 2}) of ``no reset'' \textsc{r-kfac} rapidly increase past the steady-state error of \textsc{b-kfac}, despite the behaviour in error metric (1). Thus, the errors we are most interested in are much more favorable when using \textsc{b-}updates than when using no updates. For example, at epoch 15, it took  only 10-15 skipped inverse-updates after an \textsc{r-kfac} update for the step-error to exceed the steady-state level of \textsc{b-kfac}.

\vspace{+0.5ex}
\noindent \textbf{B-KFAC vs B-R-KFAC: Relationship to \textit{Proposition 3.2}.} \textit{Figures 1} and \textit{2} show that when adding periodic \textsc{rsvd} overwritings to a \textsc{b-kfac} algorithm, the error is better for all iterations (compare plots of \textsc{b-kfac} with plots of \textsc{b-r-kfac}). This aligns with the intuition we have developed from \textit{Proposition 3.2.} 

\vspace{+0.5ex}
\noindent \textbf{B-KFAC vs B-R-KFAC vs B-KFAC-C.} The error of \textsc{b-kfac-c} lies in between the one of (the more expensive) \textsc{b-r-kfac} and (the cheaper) \textsc{b-kfac}. Thus, \textsc{b-kfac-c} allows us to trade between CPU time and error by tuning $\phi_{corct}$.

\vspace{+0.5ex}
\noindent \textbf{Average Error in Relation to Time-per-Epoch ($t_{\text{epoch}}$).} It is instructive to consider the relationship between average error and $t_{\text{epoch}}$. \textit{Table 1} summarizes the results in \textit{Figures 1} and \textit{2}, but also shows $t_{\text{epoch}}$ measurements. The average-error order obviously carries on from the one observable in \textit{Figures 1} and \textit{2}, while the $t_{\text{epoch}}$ ordering is what one would expect based on our discussions in \textit{Section 3}. Note that when considering  $t_{\text{epoch}}$, the fair comparison between \textsc{b-kfac} and \textsc{r-kfac} is when the latter has $T_{\text{inv}}=10$ (both perform an inverse-update every 10 steps). In this case, we see that \textsc{b-kfac} is much cheaper than \textsc{r-kfac}. However, increasing $T_{\text{inv}}$ to $T_{\text{inv}}=50$ gets \textsc{r-kfac} slightly cheaper per epoch than \textsc{b-kfac}.

No algorithm has all metrics better than any other. Thus, we cannot say which one will give better optimization performance. However, we can see that \textsc{r-kfac} has significantly smaller $t_{\text{epoch}}$ than \textsc{k-fac} with only marginally larger error, suggesting \textsc{r-kfac} will most likely perform better in practice (reconciling with findings in \cite{randomized_KFACS}). We can also see we dramatically reduce error of \textsc{r-kfac} ($T_{\text{inv}}=50$) by adding in \textsc{b}-updates (getting \textsc{b-r-kfac}), while the $t_{\text{epoch}}$ overhead is minimal (reconciles with our discussion in \textit{Section 3}). Finally, we see some of \textsc{b-r-kfac}'s accuracy can be given away in exchange for slightly smaller $t_{\text{epoch}}$ by turning the \textsc{rsvd}-overwriting either into a correction (getting \textsc{b-kfac-c}), or into a \textsc{b}-update (getting \textsc{b-kfac}). Note that changes in $t_{\text{epoch}}$ are relatively small when taking out/putting in \textsc{b}-updates, because these updates are applied only to the first FC layer, but  $t_{\text{epoch}}$ measures computations across all the layers.

\begin{table}[t]
	\centering
	\caption{Avg.\ Error Metrics (of each error curve at epoch 30) and $t_{\text{epoch}}$ (mean $\pm$ SD).}
	\begin{tabular}{|c|c|c|c|c|c|}
		\hline
		\makecell{Optimizer} & \makecell{Avg.\ Err.\ \\ metric 1} &\makecell{Avg.\ Err.\ \\ metric 2} & \makecell{Avg.\ Err.\ \\ metric 3} &  \makecell{Avg.\ Err.\ \\ metric 4}  &  \makecell{$t_{\text{epoch}}$ (s)} \\
		\hline
		
		\hline
		
		\hline
		
		\textsc{k-fac} $T_{\text{inv}}=50$ & 1.51e-03 & 8.42e-02 &  2.69e-01 & 14.4e-02 & $169\pm 5.6$\\
		\hline

		\textsc{r-kfac $T_{\text{inv}}=50$} &  4.65e-03 & 10.4e-02 & 2.96e-01& 14.6e-02 & $58.5\pm 0.6$\\
		\hline
		
		\textsc{r-kfac $T_{\text{inv}}=10$} &4.8e-03 & 4.9e-02 & 0.43e-01 & 0.02e-02 & $71.2\pm 1.0$\\
		\hline
		
		\textsc{b-kfac}  \ & 48.6e-03 & 1.02e-02 & 2.40e-01 & 29.0e-02 & $ 59.2 \pm 0.6$ \\
		\hline
		
		\textsc{b-kfac-c}  \ & 11.9e-03 & 1.04e-02 & 1.22e-01& 0.77e-02 & $ 59.4 \pm 0.6$ \\
		\hline

		\textsc{b-r-kfac}  \ & 3.40e-03 & 1.81e-02 & 0.40e-01 & 0.74e-02 & $59.5\pm 0.6$ \\
		\hline	
		
	\end{tabular}	

	
	\vspace{-1ex}

\end{table}

\section{Proposing a K-factor Inverse Application which is Linear in $d_M$}
So far, we have seen that for certain layers we can use the B-update to obtain low-rank inverse representations of K-factors in linear time\footnote{Compare to Cubic time in $d_M$ as is typically done in standard K-FAC \cite{KFAC}, \cite{convolutional_KFAC}, or with quadratic time in $d_M$ of Randomized-KFACs \textit{Algorithm \ref{R_KFAC_algorithm}} (see \cite{randomized_KFACS}).} in $d_M$ (though with no error guarantee). To reap the most benefits, we would like the inverse application itself to also scale no worse than linear in $d_M$. With the inverse application procedure proposed in \textit{Algorithm \ref{R_KFAC_algorithm}}, we saw we could make it quadratic. In this section, we argue that, for the layers where applying the B-update makes sense, we can make our inverse application linear in $d_M$ as well. 

However, in our numerical experiments we did not implement this feature, and left this as future work!

The idea behind our approach is simple: the gradient of each layer (in matrix form, $\text{Mat}(g^{(l)})$) is a of the form (at iteration $k$)
\begin{equation}
\text{Mat}(g_k^{(l)}) = G^{(l)}_k [A_k^{(l)}]^T,
\end{equation} 
where $G^{(l)}_k\in \mathbb R^{d_{\Gamma}^{(l)}\times n_M^{(l)}}$ and $G^{(l)}_k\in \mathbb R^{d_{\mathcal A}^{(l)}\times n_M^{(l)}}$ are the same matrices as the ones used to generate the EA K-factors as (review \textit{Section 1.5}).

Thus, we see that whenever the B-update is applicable ($n_M^{(l)}< n_{BS} + d_{\mathcal A}^{(l)}$ for the ``A'' K-factors, and $n_M^{(l)}< n_{BS} + d_{\Gamma}^{(l)}$ for the ``G'' K-factors), we can make further computational savings by avoiding the multiplication of $ G^{(l)}_k [A_k^{(l)}]^T$ in the backward pass and applying the inverse representation by first taking a product with $ G^{(l)}_k$, and then with $A_k^{(l)}$. That is, we compute the product $[\bar \Gamma^{(l)}]^{-1}\text{Mat}(g_k^{(l)}) [\bar {\mathcal A}^{(l)}_k]^{-1}$ as 
\begin{equation}
[\bar \Gamma^{(l)}_k]^{-1}\text{Mat}(g_k^{(l)}) [\bar {\mathcal A}^{(l)}_k]^{-1} = \biggl([\bar \Gamma_k^{(l)}]^{-1}G^{(l)}_k\biggr) \biggl([A_k^{(l)}]^T[\bar {\mathcal A}^{(l)}_k]^{-1}\biggr),
\end{equation}
but of course we use our low-rank inverse representations for $[\bar \Gamma_k^{(l)}]^{-1}$ and $[\bar {\mathcal A}^{(l)}_k]^{-1}$ rather than the standard EVD inverses. \textit{Algorithm \ref{linear_inverse_applicaitoN_algorithm_B_KFAC}} shows how this  works in practice.

\begin{algorithm}[H]
	\label{linear_inverse_applicaitoN_algorithm_B_KFAC}
	\caption{Linear Inverse Application: works whenever (\textsc{b-update}) can be applied}
	\tcp{The low-rank inverse representation of $[\bar \Gamma^{(l)}]^{-1}$ and $[\bar {\mathcal A}^{(l)}_k]^{-1}$ are required}
	
	Modify the back-prop\footnote{While we do have access to the matrices $G^{(l)}_k$ and $A_k^{(l)}$ in a standard K-FAC algorithm with typical back-prop, this is done through a backward and forward-pass hook. By using only using the hooks, and not modifying the back-prop, we would still compute the unnecessary quantities $\{G^{(l)}_k [A_k^{(l)}]^T\}_{l,k}$. We thus need an actual modification of the back-prop to also save on this wasteful. Using the hooks with no back-prop modification would still provide the inverse application savings discussed in this section, but it would not be the best way of implementing (would miss on an extra saving in the back-prop).} to return $G^{(l)}_k$, $A_k^{(l)}$ rather than $\text{Mat}(g_k^{(l)})$
	
	Replace the \textit{lines 8-10} of \textit{Algorithm \ref{R_KFAC_algorithm}} with: \tcp{Apply the low-rank inverse representation with linear time in $d_{\Gamma}$ and $d_{\mathcal A}$}
	
	\vspace{+2ex}
	\tcp{Estimate $[A_k^{(l)}]^T[\bar {\mathcal A}^{(l)}_k]^{-1}$, use the lowrank  represent.\ of $\bar {\mathcal A}^{(l)}_k$:}
	$[\mathbf{A}^{(l)}_k]^T := [A^{(l)}_k]^T \tilde V^{(l)}_{A, k}\big[( \tilde D^{(l)}_{A,k} + \lambda I)^{-1} - \frac{1}{\lambda}I\big] (\tilde V^{(l)}_{A,k})^T+ \frac{1}{\lambda}[A^{(l)}_k]^T$ \tcp*{$\mathcal O (rd_{\mathcal A}^{(l)}n_M^{(l)})$}
	
	\vspace{+2ex}
	\tcp{Estimate $[\bar \Gamma^{(l)}_k]^{-1}G^{(l)}_k$, use the lowrank represent.\ of $\bar \Gamma^{(l)}$:}
	$\mathbf G_k^{(l)} := \tilde V^{(l)}_{\Gamma, k}\biggl[( \tilde D^{(l)}_{\Gamma,k} + \lambda I)^{-1} - \frac{1}{\lambda}I\biggr] (\tilde V^{(l)}_{\Gamma,k})^T G^{(l)}_k + \frac{1}{\lambda}G^{(l)}_k$ \tcp*{$\mathcal O (rd_{\Gamma}^{(l)} n_M^{(l)})$}
	
	\vspace{+2ex}
	\tcp{Construct the Precond.\ Step from Partial Quantities\footnote{Note that the matrices $[\mathbf{A}^{(l)}_k]^T \approx [A_k^{(l)}]^T[\bar {\mathcal A}^{(l)}_k]^{-1}$ and $\mathbf G_k^{(l)}\approx [\bar \Gamma_k^{(l)}]^{-1}G^{(l)}_k$ are the preconditioned $[A^{(l)}_k]^T$ and $G^{(l)}_k$, and they are first defined in \textit{Algorithm \ref{linear_inverse_applicaitoN_algorithm_B_KFAC}}. }:}
	$S^{(l)}_k = \mathbf G_k^{(l)} [\mathbf{A}^{(l)}_k]^T$  \tcp{$S^{(l)}_k$ is the step for layer $l$ in matrix form}
	
\end{algorithm}

\vspace{+3ex}

We thus see that whenever the B-update is applicable ($n_M^{(l)}< n_{BS} + d_{\mathcal A}^{(l)}$ for the ``A'' K-factors, and $n_M^{(l)}< n_{BS} + d_{\Gamma}^{(l)}$ for the ``G'' K-factors), we reduce the time scaling of our inverse application from $\mathcal O(r([d_{\Gamma}^{(l)}]^2 + [d_{\mathcal A}^{(l)}]^2 ))$ (as we had in R-KFAC, \textit{Algorithm \ref{R_KFAC_algorithm}}), to $\mathcal O(r(d_{\Gamma}^{(l)} + d_{\mathcal A}^{(l)})n_M^{(l)})$. This inverse application methodology offers: 

\begin{enumerate}
	\item An improved inverse application complexity down to linear in layer size for all K-factors;
	
	\item A concrete, practical computational saving whenever the B-update is applicable\footnote{In fact, the condition is looser: practical computational savings occur when $n_M<d_M$.} (since $n_M<d_M$ when this happens).
\end{enumerate}

\section{Numerical Results}
\subsubsection{Implementation details} 
We now compare the numerical performance of our proposed algorithms: \textsc{b-kfac}, \textsc{b-r-kfac} or \textsc{b-kfac-c}, with relevant benchmark algorithms: \textsc{r-kfac} (\textit{Algorithm 1}; \cite{randomized_KFACS}), \textsc{k-fac} (\cite{KFAC}), and \textsc{seng} (the state of art implementation of NG for DNNs; \cite{SENG}). We consider the CIFAR10 classification problem with a modified\footnote{We reduce all the pooling kernels size from 2x2 to 2x1. We do this to increase the width of the FC layer 0 of VGG16\_bn (by $32\times$), to put us in a position where K-Factor computations in the FC layers are not negligible. Thus, we have FC layer 0: 16384-in$\times$2048-out with dropout ($p=0.5$), and the final FC layer: 2048-in$\times$10-out.} version of batch-normalized VGG16 (VGG16\_bn). All experiments ran on a single \textit{NVIDIA Tesla V100-SXM2-16GB} GPU. The accuracy and loss we refer to are always on the \textit{test set}.

For \textsc{seng}, we used the \textit{official github repo} implementation with the hyperparameters\footnote{\textbf{Repo:} https://github.com/yangorwell/SENG. \textbf{Hyper-parameters:} see Appendix. } directly recommended by the authors for the problem at hand.

For all algorithms except \textsc{seng}, we use $\rho=0.95$, $n_{\text{BS}}=256$ and $T_{\text{updt}}=25$, weight decay of $7e-04$, no momentum, a clip parameter of $0.07$ and a learning rate schedule of $\alpha_k = 0.3 - 0.1\mathbb I_{n_{ce}(k)\geq 2} - 0.1\mathbb I_{n_{ce}(k)\geq 3} - 0.07\mathbb I_{n_{ce}(k)\geq 13} - 0.02\mathbb I_{n_{ce}(k)\geq 18} - 0.007\mathbb I_{n_{ce}(k)\geq 27} - 0.002\mathbb I_{n_{ce}(k)\geq 40}$ (where $n_{\text{ce}}(k)$ is the number of the current epoch at iteration $k$). For all these algorithms, we set the regularization to be depend on layer, K-factor type ($\mathcal A$ vs $\Gamma$), and iteration as\footnote{$\mathcal M$ can be $\mathcal A$ or $\Gamma$ - see \textit{lines 16-17} of \textit{Algorithm 1}.  $\lambda^{(\mathcal M)}_{\text{max},k,l}$ is the maximum eigenvalue of our possibly approximate representation of the K-factor $\mathcal M$ at layer $l$, iteration $k$.} $\lambda_{k,l}^{(\mathcal M)}=\lambda_{\text{max},k,l}^{(\mathcal M)}\phi_{\lambda,k}$  with the schedule $\phi_{\lambda,k} = 0.1 - 0.05\mathbb I_{n_{\text{ce}}(k)\geq 25} - 0.04\mathbb I_{n_{\text{ce}}(k)\geq 35}$. 

For \textsc{k-fac} and \textsc{r-kfac} we set $T_{\text{inv}}=250$. We also consider an \textsc{r-kfac} which uses all the previous settings but $T_{\text{inv}}=25$ - we refer to it ``\textsc{r-kfac} $T_{\text{inv}}=25$''. The hyperparameters specific to \textsc{r-kfac} were set to $n_{\text{pwr-it}}= 4$, target-rank schedule $r(k) = 220 + 10\mathbb I_{n_{\text{ce}(k)}\geq 15}$, and oversampling parameter schedule $r_o(k) = 10 + \mathbb I_{n_{ce}(k)\geq 22}  + \mathbb I_{n_{\text{ce}(k)}\geq 30}$ (see \cite{randomized_KFACS} for details).

For \textsc{b-kfac}, \textsc{b-r-kfac} and \textsc{b-kfac-c} we set the \textit{truncation rank schedule} to be the same as the {target-rank schedule} ($r(k)$) of \textsc{r-kfac}. For \textsc{b-kfac} we used $T_{\text{Brand}}=125$. \textsc{b-kfac-c} had $T_{\text{Brand}}=125$, $T_{corct} = 500$, and $\phi_{corct} = 0.5$. \textsc{b-r-kfac} had $T_{\text{Brand}}=25$, and $T_{\text{RSVD}}=250$.

Recall from \textit{Section 3.5} that the implemented \textsc{b-kfac}, \textsc{b-r-kfac} and \textsc{b-kfac-c} use the corresponding proposed updates only for the first FC layer, and \textsc{r-kfac} updates (\textit{Algorithm 1}; \cite{randomized_KFACS}) for all other layers. The hyperparameters of the \textsc{r-kfac} part of \textsc{b-kfac}, \textsc{b-r-kfac} or \textsc{b-kfac-c} are the ones we described above.

\subsection{Algorithms Optimization Performance Comparison}
\textit{Table 2} shows a summary of results. We make the following observations:

\textbf{Benchmark 0: \textsc{seng}.} Relatively large number of epochs to target accuracy but very low $t_{\text{epoch}}$, giving the best performance across all benchmarks (and all algorithms in general) for all target test accuracies apart from $91\%$.

\textbf{Benchmark 1: \textsc{k-fac}.} the weakest of all algorithms with very large $t_{\text{epoch}}$, and surprisingly (unknown reason), much larger no.\ of epochs to target accuracy than any of its sped-up versions (all of which approximate the K-factors). 

\textbf{Benchmark 2: \textsc{r-kfac}.} Moderate number of epochs to certain accuracy and moderate $t_{\text{epoch}}$, giving a performance always better than \textsc{k-fac} (benchmark 1). Nevertheless, it only outperforms \textsc{seng} (benchmark 0) for $91\%$ test accuracy.

\textbf{\textsc{b-kfac}:} Has the lowest computational cost per epoch ($t_{\text{epoch}}$) across all \textsc{kfac}-based algorithms, while taking nearly the same amount of epochs to a target test accuracy as the other \textsc{kfac}-based algorithms. It outperforms \textsc{k-fac} and \textsc{r-kfac} (benchmarks 1 and 2) for all considered target accuracies, being \textit{the best performing \textsc{b-}$(\cdot)$ variant}. It outperforms \textsc{seng} only for $91\%$ test accuracy.

\textbf{\textsc{b-r-kfac}:} essentially upgrades \textsc{r-kfac} (with $T_{inv}=250$) to also perform a \textsc{b}-update every time new K-factor information arrives. This improves $\mathcal N_{\text{acc}\geq 93\%}$ but makes $t_{\text{epoch}}$ worse. Over-all it seems that the extra accuracy gained through introducing \textsc{b}-updates is favourable for large target test accuracy: \textsc{b-r-kfac} reaches $93.5\%$ acc.\ 8 times while \textsc{r-kfac} ($T_{\text{inv}}=250$ never does so).

\textbf{\textsc{b-kfac-c}:} Lies in between \textsc{b-kfac} and \textsc{b-r-kfac} in terms of both $t_{\text{epoch}}$ and $\mathcal N_{\text{acc}\geq93\%}$, but provids a worse cost-accuracy trade-off than either of these in this case. Nevertheless, it outperforms \textsc{r-kfac} and \textsc{k-fac} benchmarks for all considered target test accuracies, and outperforms \textsc{seng} for low target accuracy.

\begin{table}[t]
	\centering
	\caption{Optimizer Performance Results Summary. We perform 10 runs of 50 epochs for each considered solver. Columns 2-4 show times to a target test accuracy. Columns 5 shows the time per epoch. Column 6 shows how many runs get to $93.5\%$ test accuracy. All solvers get to $93\%$ accuracy 10 out of 10 times. The final column shows the number of epochs to $93\%$ test accuracy. All results concerning times and number of epochs are shown as empirical mean $\pm$ empirical st.\ dev. The results for columns 2,3,7 use 10 samples (since 10/10 runs get to $93\%$ acc.). The results in column 5 use only the runs which got to $93.5\%$. The results in column 5 use $500$ samples. All times are in seconds.}
	\begin{tabular}{|c|c|c|c|c|c|c|c|c|c|}
		\hline
		\makecell{} & \makecell{ $t_{acc\geq91\%}$} &\makecell{$t_{acc\geq93\%}$} & \makecell{$t_{acc\geq93.5\%}$} &  \makecell{$t_{\text{epoch}}$}  &  \makecell{ $\#$ hit $93.5\%$} & \makecell{$\mathcal N_{\text{acc}\geq 93\%}$} \\
		\hline
		
		\hline
		
		\hline
		
		\textsc{seng} &  $999.0\pm54.1$ & $1098,\pm37.5$ & $1144\pm51$ & $25.4\pm0.81$ & 10 in 10 &  $43.1\pm1.5$ \\
		\hline
		
		\textsc{k-fac} & $2610\pm 213$ & $4021\pm 433$ & N/A & $97.8\pm 8.2$ & 0 in 10 & $41.1\pm4.6$\\
		\hline
		
		
		\textsc{r-kfac} &  $920.3\pm24.5$ & $1357\pm 38.7$ & N/A & $47.7\pm0.54$ & 0 in 10 & $28.9\pm0.8$\\
		\hline
		
		\makecell{\textsc{r-kfac}\\ $T_{\text{inv}}=25$} &  $1019\pm3.05$ & $1526\pm49.3$ & $2108\pm373$ & $53.3\pm0.69$ & 6 in 10 & $28.5\pm0.9$\\
		\hline
		
		\textsc{b-kfac}  \ & $894.0\pm 22.9$ & $1325\pm 58.9$ & $1913\pm 125$ & $46.3\pm0.62$ & 10 in 10 & $28.6\pm 1.2$ \\
		\hline
		
		\textsc{b-kfac-c}  \ & $911.2\pm 18.9$ & $1352.3\pm 32.54$ & $2031\pm 312$ & $47.4\pm0.45$ & 6 in 10 & $28.5\pm0.7$ \\
		\hline

		\textsc{b-r-kfac}  \ & $945.9\pm 39.6$ & $1324.8\pm 58.9$ & $1975\pm 323$ & $48.4\pm0.53$ & 8 in 10 & $28.3\pm 0.5$ \\
		\hline

	\end{tabular}	
	
\end{table}

\section{Conclusion}
By exploiting the EA construction paradigm of the K-factors, we proposed an online inverse-update to speed-up \textsc{k-fac} (\cite{KFAC}) for FC layers. If we use the update exclusively, we obtian the K-factor inverse representation in linear time scaling w.r.t.\ layer size (as opposed to quadratic for \textsc{r-kfac}s \cite{randomized_KFACS}, and cubic for standard \textsc{k-fac} \cite{KFAC}, \cite{convolutional_KFAC}). This update relied on Brand's algorithm (\cite{Brand2006}), and we called it the ``\textsc{b}-update'' (of K-factors inverses). We saw the update is useful only when $d_M>n_M + r$, which typically holds for FC layers.

In these cases, we saw the \textsc{b}-update is exact, but only remains cheap if we constrain our approximate K-Factors representation to be (very) low-rank - which we practically achieved through an SVD-optimal rank-$r$ truncation just before each \textsc{b}-update.  We argued that based on results presented in \cite{randomized_KFACS}, (EA) K-Factors typically have significant eigenspectrum decay, and thus a very low-rank approximation for them would actually have low error. 

We also saw that whenever we can apply the B-update, our inverse application technique can be improved to reduce time scaling from quadratic\footnote{Or cubic for standard \textsc{k-fac}.} in layer size (as for \textsc{rs-kfac}) down to linear. We did not implement this inverse application methodology in this paper however (this is future work).

The \textsc{b-}update, together with the truncation, and the proposed inverse application technique gave \textsc{b-kfac}. The algorithm (\textsc{b-kfac}) is an approximate \textsc{k-fac} implementation for which the preconditioning part scales (over-all) linearly in layer size. Notably however, ``pure'' \textsc{b-kfac} is only applicable to some layers (and we have to use \textsc{rs-kfac} for the others). Compared to quadratic scaling in layer size for \textsc{rs-kfac} (\cite{RSVD_paper}) or cubic for \textsc{k-fac} (\cite{KFAC}, \cite{convolutional_KFAC}), the improvement proposed here is a substantial improvement. Though there is no error guarantee bounding the \textsc{b-kfac} preconditioning error, we saw with numerical case-studies which revealed the \textsc{b-kfac} error was acceptable.

We saw that the \textsc{b-}update, other than being used alone to give \textsc{b-kfac}, can also be combined with updates like \textsc{rs-kfac} updates (\textsc{rsvd} updates) to give different algorithms with different empirical properties. By comparing \textsc{b-kfac} with \textsc{r-kfac} (\textsc{rs-kfac} in \cite{randomized_KFACS}), we noted that we may be able to increase the K-Factor representation accuracy of \textsc{b-kfac} by adding in periodic \textsc{rsvd}\textit{ ``overwritings''}, which gave the \textsc{b-r-kfac} algorithm. We saw that the \textsc{b-r-kfac} can also be seen as an \textsc{r-kfac} algorithm to which we introduce \textsc{b-}updates at times when no \textsc{rsvd} would have been performed, with the aim of better controlling the K-factor representation error, at minimal cost. We also noted we may change the \textsc{rsvd}-overwriting with a cheaper but less accurate \textit{``correction''}, in order to obtain customizable time-accuracy trade-offs, giving \textsc{b-kfac-c}.

Numerical results concerning K-Factors errors show that our all our proposed algorithms (\textsc{b-kfac}, \textsc{b-r-kfac}, and \textsc{b-kfac-c}) had errors comparable to \textsc{k-fac} (\cite{KFAC}) while offering an $\approx3\times$ speed-up per epoch. W.r.t.\ the more competitive \textsc{r-kfac} (\cite{randomized_KFACS}), our proposed algorithms offered similar metrics but more trade-offs to choose from. Notably, \textsc{b-r-kfac} was significantly better than \textsc{r-kfac} - across all investigated error metrics at minimal computational overhead. Numerical results concerning optimization performance show \textsc{b-kfac} and \textsc{b-kfac-c} consistently outperform \textsc{r-kfac} (the best \textsc{k-fac} benchmark; \cite{randomized_KFACS}) by a \textit{moderate amount}, while \textsc{b-r-kfac} only does so for relatively large target test accuracy. All our \textsc{b-}$(\cdot)$ algorithms outperform \textsc{seng} (the state of art; \cite{SENG}) for low target test accuracy. 

\vspace{+2ex}
\noindent \textbf{Future work} involves implementing the proposed inverse application methodology and re-running numerical experiments.

\subsubsection{Acknowledgments}

Thanks to \textit{Jaroslav Fowkes} and \textit{Yuji Nakatsukasa} for useful discussions. I am funded by the EPSRC CDT in InFoMM (EP/L015803/1) together with Numerical Algorithms Group and St.\ Anne's College (Oxford).

\end{document}


\pagenumbering{arabic}
\renewcommand*{\thepage}{S\arabic{page}}
\title{Supplementary Material: Brand New KFACs}

\author{Constantin Octavian Puiu\orcidID{0000-0002-1724-4533} \Letter}

\authorrunning{C. O. Puiu}
%
\institute{University of Oxford, Mathematical Institute,\\
	\email{constantin.puiu@maths.ox.ac.uk}\\
}

\toctitle{Brand New KFACs}
\tocauthor{Constantin Octavian Puiu}

\maketitle              
\newtheorem*{Proposition_3_2}{Proposition 3.2: Pure \textsc{b-kfac} vs over-writing $\mathcal B_i = \tilde {\mathcal M}_{R;i,r}$ exactly once}

\newtheorem*{Proposition_4_1}{Proposition 4.1: Error of Doing nothing vs Error of B-updates}
\newtheorem*{Proposition_4_2}{Proposition 4.2: $\norm{E_j}_F$ ($j\geq 1$) upper bounds for No update vs for B-update}

\appendix

\renewcommand{\theequation}{S.\arabic{equation}}

\setcounter{equation}{0}

\section{Proof of Proposition 3.2}
We reiterate \textit{Proposition 3.2} for convenience. We also reiterate the equations of the K-factors related processes we are interested in.

Consider an arbitrary EA K-Factor $\mathcal M_k$ (may be either $\bar{\mathcal A}_k^{(l)}$ or $\bar{\Gamma}_k^{(l)}$ for any $l$) where we have incoming (random) updates $M_kM_k^T$ with $M_k\in\mathbb R^{n\times n_{\text{BS}}}$ at iteration $i$. This follows the process
\begin{equation}
\begin{split}
\mathcal M_0 = M_0M_0^T,\,\,\,
\mathcal M_{j}= \rho \mathcal M_{j-1} + (1-\rho)M_jM_j^T \,\,\forall j\geq 1,
\end{split}
\label{eqn_the_KFAC_process}
\end{equation}
and can alternatively be written as 
\begin{equation}
\mathcal M_k = \rho^k \mathcal M_0 + (1-\rho)\sum_{j=1}^k\rho^{k-j}M_jM_j^T = \sum_{i=0}^k\kappa(i)\rho^{k-i}M_iM_i^T\,\,\,\forall k\geq 0,
\label{true_Kfactor_0} 
\end{equation}
where $\kappa(i) := 1-\rho\mathbb I_{i>0}$. Ignoring the projection error of \textsc{rsvd} (which is very small for our purpose [3], when performing \textsc{rs-kfac} (with target rank $r$) instead of \textsc{k-fac} we effectively estimate $\mathcal M_k$ as 
\begin{equation}
\begin{split}
&\tilde {\mathcal M}_{R;k,r} = U_{\mathcal M_k,r}  U_{\mathcal M_k,r}^T\mathcal M_kU_{\mathcal M_k,r}U_{\mathcal M_k,r}^T\,\,\,\forall k\geq 0,\,\,\,\text{where}
\\
U_{\mathcal M_k} D_{\mathcal M_k}  U_{\mathcal M_k}^T \,\,&\SVDeq\,\,\mathcal M_k =  \sum_{i=0}^k\kappa(i)\rho^{k-i}M_iM_i^T, \,\,\,\text{and}\,\,\, U_{\mathcal M_k,r} := U_{\mathcal M_k}[:,:r].
\end{split}
\label{eqn_the_R_process}
\end{equation}
Conversely, \textsc{b-kfac} effectively estimates $\mathcal M_k$ as $\tilde {\mathcal M}_{B,k}$, where $\tilde {\mathcal M}_{B,k}$ is given by
\begin{equation}
\begin{split}
\tilde {\mathcal M}_{B,i+1}:=\rho  U_{\tilde {\mathcal M}_{B,i},r}&U_{\tilde {\mathcal M}_{B,i},r}^T\tilde {\mathcal M}_{B,i}U_{\tilde {\mathcal M}_{B,i},r}U_{\tilde {\mathcal M}_{B,i},r}^T+(1-\rho)M_{i+1}M_{i+1}^T\,\forall i\geq 0,
\\
\text{with}\,\tilde {\mathcal M}_{B,0} = M_0M_0^T,\,\,\,&U_{\tilde {\mathcal M}_{B,i}}D_{\tilde {\mathcal M}_{B,i}}U_{\tilde {\mathcal M}_{B,i}}^T\SVDeq\tilde {\mathcal M}_{B,i},\,\, U_{\tilde {\mathcal M}_{B,i},r}:=U_{\tilde {\mathcal M}_{B,i}}[:,:r];
\\
\text{we also define }& \mathcal B_i := U_{\tilde {\mathcal M}_{B,i},r}U_{\tilde {\mathcal M}_{B,i},r}^T\tilde {\mathcal M}_{B,i}U_{\tilde {\mathcal M}_{B,i},r}U_{\tilde {\mathcal M}_{B,i},r}^T\,\,\,\forall i\geq 0.
\end{split}
\label{eqn_the_B_process}
\end{equation}
\begin{Proposition_3_2}
	For $j\geq 1$, let $\tilde {\mathcal M}^{R@i}_{i+j}$ and $\mathcal B^{R@i}_{i+j}$ be the $\tilde {\mathcal M}_{i+j}$ and $\mathcal B_{i+j}$ produced by process (\ref{eqn_the_B_process}) after over-writing $\mathcal B_i = \tilde {\mathcal M}_{R,i,r}$ at $i\geq0$. The error when doing so ($\forall m\geq 1$) is
	\begin{equation}
	\begin{split}
	E^{\text{R@i}}_{i+m}:= (\mathcal M_{i+m} - \tilde {\mathcal M}^{R@i}_{B,i+m}) = \rho^{m}(\mathcal M_i - \tilde {\mathcal M}_{R,i,r})
	+\sum_{j=1}^{m-1}\rho^{m-j}(\tilde{\mathcal M}^{R@i}_{B,i+j}-\mathcal B^{R@i}_{i+j}).
	\end{split}
	\label{B_overwritten_err}
	\end{equation}
	When performing pure \textsc{b-kfac} we have the error at each $i+m$  ($\forall m\geq 1$) as
	\begin{equation}
	\begin{split}
	E^\text{(pure-B)}_{i+m}:= (\mathcal M_{i+m} - \tilde {\mathcal M}_{B,i+m}) = \rho^{m}(\mathcal M_i - \mathcal B_{i})
	+\sum_{j=1}^{m-1}\rho^{m-j}(\tilde{\mathcal M}_{B,i+j}-\mathcal B_{i+j}).
	\end{split}
	\label{B_pure_err}
	\end{equation}
	Further, all the quantities within $(\cdot)$ are sym.\ p.s.d.\ matrices for any index $\geq 0$.
	
\end{Proposition_3_2}
\textit{Proof.} \textit{\textbf{Part 1:} Prove equations form.} We prove (\ref{B_overwritten_err}) and (\ref{B_pure_err}) simultaneously noting that both the overwritten \textsc{b-kfac} process $\{\tilde {\mathcal M}^{R@i}_{B,i+m}\}_{m\geq 1}$ and the pure \textsc{b-kfac} process $\{\tilde {\mathcal M}_{B,i+m}\}_{m\geq 1}$ evolve in the same way for $m\geq 1$, the only difference being the initial condition at $i$. Thus, let $X_i$ be the initial condition for $\{\tilde {\mathcal M}_{B,i+m}\}_{m\geq 1}$. By performing the derivation for arbtrary $X_i$, and at the end setting $X_i\leftarrow \tilde {\mathcal M}_{R,i,r}$ for the \textsc{rsvd}-\textit{overwritten} \textsc{b-kfac} process (at iteration $i$)  and $X_i\leftarrow \tilde {\mathcal M}_{B,i}$ for pure \textsc{b-kfac} process we get our desired equations in each case (also noting that different initial conditions mean all subsequent iterates are in principle different - even though the evolution follows the same law).

Using equations (\ref{eqn_the_KFAC_process}) and (\ref{eqn_the_B_process}) we have
\begin{equation}
\mathcal M_{i+m} - \tilde {\mathcal M}_{B,i+m} = \rho(\mathcal M_{i+m-1} - \mathcal B_{i+m-1}),
\label{eqn_7_appendix_prop_3_2}
\end{equation}
\begin{equation}
\mathcal M_{i+m} - \tilde {\mathcal M}_{B,i+m} = \rho(\mathcal M_{i+m-1} - \tilde {\mathcal M}_{B,i+m-1}) + \rho(\tilde {\mathcal M}_{B,i+m-1} - \mathcal B_{i+m-1})
\label{eqn_apply_this_recursively_onto_itself_prop_3_2}
\end{equation}
Now, applying (\ref{eqn_apply_this_recursively_onto_itself_prop_3_2}) recursively $m-1$ (and noting our initial condition was $\tilde {\mathcal M}_{B,i} = X_i$), we get
\begin{equation}
\mathcal M_{j+m} - \tilde {\mathcal M}_{j+m} = \rho^{m}(\mathcal M_i - X_{i}) + \sum_{j=1}^{m-1}\rho^{m-j}(\tilde{\mathcal M}_{B,i+j}-\mathcal B_{i+j}).
\end{equation}
Now imposing the initial condition of the \textsc{rsvd}-\textit{overwritten} \textsc{b-kfac} process, $X_i\leftarrow \tilde {\mathcal M}_{R,i,r}$ gives (\ref{B_overwritten_err}). Conversely, imposing the initial condition of the pure \textsc{b-kfac} process, $X_i\leftarrow \tilde {\mathcal M}_{B,i}$ gives (\ref{B_pure_err}). Note that the quantities in equation (\ref{B_overwritten_err}) have ``$R@i$'' superscripts to highlight that the two processes are different (since the intial condition was different - even though the evolution law is the same).

\textit{\textbf{Part 2:} Prove all matrices in $(\cdot)$ are symmetric positive semi-definite (s.p.s.d.).} For a p.s.d.\ matrix $A$, we will use the standard notation $A \succcurlyeq 0$.

To prove $\mathcal M_k - \tilde {\mathcal M}_{R,i,r}\succcurlyeq 0 $, note that when the \textsc{rsvd} projection error is zero\footnote{We neglect it for our purpose as it is very small for our case in practice.}, $\tilde {\mathcal M}_{R,i,r}$ represents the first $r$ eigenmodes of $\mathcal M_k$. Thus, the eigenvalues of $\mathcal M_k - \tilde {\mathcal M}_{R,i,r}$ are the smallest $n-r$ eigenvalues of $\mathcal M_k$. But $\mathcal M_k$ is s.p.s.d., so $\mathcal M_k - \tilde {\mathcal M}_{R,i,r}\succcurlyeq 0$ $\forall i\geq 0$.

To prove $(\tilde{\mathcal M}_{B,i+j}-\mathcal B_{i+j})\succcurlyeq 0$ and $(\tilde{\mathcal M}^{R@i}_{B,i+j}-\mathcal B^{R@i}_{i+j})\succcurlyeq 0$ use the definition of $B_i$ in (\ref{B_overwritten_err}) and note that both these matrices are truncation errors. Since $\mathcal B_{i+j}$ is the first $r$ eigenmodes of $\tilde{\mathcal M}_{B,i+j}$ the results immediately follow using the same argument as in the paragraph above.

To prove $\mathcal M_i - \mathcal B_i \succcurlyeq 0$, note that 
\begin{equation}
\mathcal M_i - \mathcal B_i = (\mathcal M_{i} - \tilde {\mathcal M}_{B,i}) +(\tilde {\mathcal M}_{B,i} - \mathcal B_{i}) \,\,\,\forall i\geq 0
\end{equation}
Now, using (\ref{eqn_7_appendix_prop_3_2}), for  $\mathcal M_{i} - \tilde {\mathcal M}_{B,i}$
\begin{equation}
\mathcal M_i - \mathcal B_i = \rho(\mathcal M_{i-1} - \mathcal B_{i-1}) +(\tilde {\mathcal M}_{B,i} - \mathcal B_{i}) \,\,\,\forall i\geq 1
\label{the_other_recurssion_similar_to_8}
\end{equation}
Applying (\ref{the_other_recurssion_similar_to_8}) recursively $i-1$ times we get 
\begin{equation}
\mathcal M_i - \mathcal B_i  = \rho^{i}(\mathcal M_0 - \mathcal B_0)+ \sum_{q=1}^i\rho^{i-q}(\tilde {\mathcal M}_{B,q}-\mathcal B_q) \,\,\,\forall i\geq 0
\end{equation}
All the terms $(\tilde {\mathcal M}_{B,q}-\mathcal B_q)\succcurlyeq 0 $ by the argument above (they are rank-$r$ truncation errors). Further, we have $\mathcal M_0 = \tilde{\mathcal M}_{B,0}= M_0M_0^T$ by our choice, and thus $\mathcal M_0 - \mathcal B_0 = \tilde{\mathcal M}_{B,0} - \mathcal B_0\succcurlyeq 0$ by the exact same argument. A positively-weighted sum of p.s.d.\ matrices is p.s.d, so $\mathcal M_i - \mathcal B_i\succcurlyeq 0$ $\forall i\geq 0$.

We have now proved that the r.h.s.\ of both (\ref{B_overwritten_err}) and (\ref{B_pure_err}) contains only p.s.d.\ matrices, so by using the positively-weighted sum argument again, the l.h.s.\ of equations (\ref{B_overwritten_err}) and (\ref{B_pure_err}) are also p.s.d.

We only need to prove symmetry of all matrices in $(\cdot)$ of equations (\ref{B_overwritten_err}) and (\ref{B_pure_err}). This is trivial by noting that the process $\{\mathcal M_i\}_i$ is symmetric and that the processes $\{\mathcal B_{i+m}\}_{m\geq 1}$, $\{\tilde {\mathcal M}_{B,i+m}\}_{m\geq 1}$ are symmetric when their initialization $X_i$ is symmetric (which is by our choice). This completes the proof. $\qed$

\section{Proof of Proposition 4.1}
We reiterate the statement of \textit{Proposition 5.1} for convenience.
\begin{Proposition_4_1}
	Let $\tilde {\mathcal M}_k$ be an approx.\ of $\mathcal M_k$ which is obtained by performing an \textsc{rs-kfac} update at $k=0$, and either no other update thereafter, or B-updates (every step) thereafter. The error in $\mathcal M_k$ when using one of these approximations is of the form
	\begin{equation}
	\mathcal M_k - \tilde {\mathcal M}_{k} = \sum_{i=0}^{k} \kappa(i)\rho^{k-i}E_i,\,\,\,\,\text{with}\,\,\,E_0 =(\mathcal M_0 - \tilde {\mathcal M}_{R;0,r}).
	\label{prop5_1_general_err_form}
	\end{equation}
	When performing RSVD initially (at $k=0$), and no update thereafter we have 
	\begin{equation}
	E_i =  M_i M_i^T - \tilde {\mathcal M}_{R;0,r}.
	\label{err_with_no_update_performed}
	\end{equation}
	When performing RSVD initially (at $k=0$), and B-updates thereafter we have
	\begin{equation}
	E_i = \frac{1}{1-\rho}(\tilde{\mathcal M}_{i} - \mathcal B_i)\,\,\,\,\forall i\in\{1,...,k-1\},\,\,\,\text{and}\,\,\,E_k = 0,
	\label{err_with_B_update_performed}
	\end{equation}
	and $\tilde{\mathcal M_i} = \tilde {\mathcal M}_{B,i}$.  where $\mathcal B_i$ is as in (\ref{eqn_the_B_process}).
\end{Proposition_4_1}
\textit{Proof.} \textit{\textbf{Part 1:} $\tilde {\mathcal M}_k$ is obtained with a \textsc{rs-kfac} update at $k=0$ and no update thereafter.}  In this case, we have $\tilde {\mathcal M}_k = \tilde {\mathcal M}_{R;,0,r}$ $\forall k\geq 0$. Using this, and (\ref{true_Kfactor_0}):
\begin{equation}
\mathcal M_k -\tilde{\mathcal M}_k = \sum_{i=0}^k\kappa(i)\rho^{k-i}M_iM_i^T - \tilde{\mathcal M}_{R;,0,r} = \sum_{i=0}^k\kappa(i)\rho^{k-i}(M_iM_i^T - \tilde{\mathcal M}_{R;,0,r}),
\end{equation}
which proves the claim about $\{E_i\}_{i\geq 0}$ when performing RSVD initially (at $k=0$), and no update thereafter.

\textit{\textbf{Part 2:}  For \textsc{rs-kfac} with an RS update at $k=0$, and B-updates at each $k$ thereafter.} In this case, we can use equation () o \textit{Prposition 3.2} directly with $i\leftarrow 0$ (we ``overwrite'' the \textsc{b-kfac} once at $k=0$ and B-update thereafter). We have 
\begin{equation}
\mathcal M_{m} - \tilde {\mathcal M}^{R@0}_{B,m} = \rho^{m}(\mathcal M_0 - \tilde {\mathcal M}_{R,0,r})
+\sum_{j=1}^{m-1}\rho^{m-j}(\tilde{\mathcal M}^{R@0}_{B,j}-\mathcal B^{R@0}_{j})
\label{eqn_before_dropping_superscripts}
\end{equation}
Now, using (\ref{eqn_the_KFAC_process}) and (\ref{eqn_the_B_process}) note that $\tilde {\mathcal M}_{B,0} = M_0M_0^T = \mathcal M_0$, which implies $U_{\tilde {\mathcal M}_{B,0},r} = U_{\mathcal M_0,r}$. This gives the important identity
\begin{equation}
\begin{split}
\tilde {\mathcal M}_{R;,0,r} =  U_{\mathcal M_0,r}  U_{\mathcal M_0,r}^T\mathcal M_0U_{\mathcal M_0,r}U_{\mathcal M_0,r}^T =\\ U_{\tilde {\mathcal M}_{B,0},r}U_{\tilde {\mathcal M}_{B,0},r}^T\tilde {\mathcal M}_{B,0}U_{\tilde {\mathcal M}_{B,0},r}U_{\tilde {\mathcal M}_{B,0},r}^T = \mathcal B_0
\end{split}
\end{equation}
Thus, the \textsc{b-kfac} algorithm overwritten exactly once at $k=0$ is just the standard \textsc{b-kfac} algorithm - so we can drop the ``$R@0$'' superscripts in (\ref{eqn_before_dropping_superscripts}) to get
\begin{equation}
\mathcal M_{m} - \tilde {\mathcal M}_{B,k} = \rho^{k}(\mathcal M_0 - \tilde {\mathcal M}_{R,0,r})
+\sum_{j=1}^{k-1}\rho^{k-j}(\tilde{\mathcal M}_{B,j}-\mathcal B_{j}).
\label{eqn_proved_B_R_kfac_superscripts_dropped}
\end{equation}
Rearranging (\ref{eqn_proved_B_R_kfac_superscripts_dropped}) we get
\begin{equation}
\mathcal M_{m} - \tilde {\mathcal M}_{B,k} = \rho^{k}(\mathcal M_0 - \tilde {\mathcal M}_{R,0,r})
+\sum_{j=1}^{k-1}\rho^{k-j}\kappa(j)\biggl[\frac{1}{1-\rho}(\tilde{\mathcal M}_{B,j}-\mathcal B_{j})\biggr].
\label{eqn_proved_B_R_kfac_superscripts_dropped_rearranged}
\end{equation}
Equation (\ref{eqn_proved_B_R_kfac_superscripts_dropped_rearranged}) proves the claim about $\{E_i\}_{i\geq 0}$ when performing RSVD initially (at $k=0$), and B-update thereafter. $\qed$
\section{Proof of Proposition 4.2}
We reiterate the statement of \textit{Proposition 5.2} for convenience.
\begin{Proposition_4_2}
	When performing RSVD initially (at $k=0$), and no updates thereafter $\norm{E_j}_F$ can get as high as
	\begin{equation}
	\norm{E_j}_F =  \sqrt{\norm{M_jM_j^T}^2_F + \norm{\tilde {\mathcal M}_{R,0,r}}^2_F}, \,\,\,\forall j\in\{1,...,k\}.
	\label{eqn_1_to_prove_in_prop_4_2}
	\end{equation}
	When performing RSVD initially, and B-updates thereafter $\norm{E_j}$ is bounded as:
	\begin{equation}
	\norm{E_j}_F\leq \norm{M_jM_j^T}_F, \,\,\,\forall j\in\{1,...,k-1\}, \,\,\,\text{and}\,\,\, E_k = 0.
	\label{eqn_2_to_prove_in_prop_4_2}
	\end{equation}
\end{Proposition_4_2}
\textit{Proof.} \textit{\textbf{Part 1:}  For \textsc{rs-kfac} with an RS update at $k=0$ and no update thereafter.}  Let $D_{\mathcal M_0,r}:=D_{\mathcal M_0}[:r,:r]$, and note that $\tilde {\mathcal M}_{R;0,r} \SVDeq U_{\mathcal M_0,r} D_{\mathcal M_0,r} U_{\mathcal M_0,r}^T$ by (\ref{eqn_the_R_process}). We have thus have
\begin{equation}
E_j =  M_j M_j^T - \tilde {\mathcal M}_{R;0,r} = M_j M_j^T - U_{\mathcal M_0,r} D_{\mathcal M_0,r} U_{\mathcal M_0,r}^T.
\label{err_with_no_update_performed_2}
\end{equation}
Let 
\begin{equation}
V_{M_j}D_{M_j}V_{M_j}^T\SVDeq M_jM_j^T
\end{equation}
Since $r+c < n$, we can have a case when $V_{M_j}^T U_{\mathcal M_0,r} = 0$. In this case, we have the SVD of $E_j$ as
\begin{equation}
E_j \SVDeq 
\begin{bmatrix}
V_{M_j} & U_{\mathcal M_0,r} & \\
\end{bmatrix}
\begin{bmatrix}
D_{M_j} & 0 & \\
0 & -D_{\mathcal M_0,r} &
\end{bmatrix}
\begin{bmatrix}
V_{M_j}^T &\\
U_{\mathcal M_0,r}^T  
\end{bmatrix}.
\end{equation}
Thus, in this case
\begin{equation}
\norm{E_j}_F = \sqrt{\norm{D_{M_j}}^2_F + \norm{\tilde D_0}^2_F}
\end{equation}
By using the fact that $\norm{UDU^T}^2_F=\norm{D}^2_F$ for arbitrary orthonormal $U$ of appropriate dimensions, and the fact that $V_{M_j}^TV_{M_j} = I$, and $U_{\mathcal M_0,r}^T U_{\mathcal M_0,r} = I$ we get
\begin{equation}
\norm{E_j}_F = \sqrt{\norm{M_jM_j^T}^2_F + \norm{ \tilde {\mathcal M}_{R;0,r}}^2_F},
\end{equation}
which completes the proof\footnote{Another proof is to use the relationship between trace and Frobenius norm directly in (), and then use further trace properties. This proof is slightly longer, but gives some further insights.} for equation (\ref{eqn_1_to_prove_in_prop_4_2}) of \textit{Proposition 4.2}.

\textit{\textbf{Part 2:}  For \textsc{rs-kfac} with an RS update at $k=0$, and B-updates at each $k$ thereafter.} From \textit{Proposition 4.1}, we have

\begin{equation}
\begin{split}
E_i = \frac{1}{1-\rho}(\tilde{\mathcal M}_{i} - \mathcal B_i) = \frac{1}{1-\rho}\text{Trc\_err}_{\text{SVD};r}\big(\tilde {\mathcal M}_{B,i}\big)
\\\,\,\forall i\in\{1,...,k-1\},
\end{split}
\label{err_with_B_update_performed_2}
\end{equation}
where $\text{Trc\_err}_{\text{SVD};r}(\cdot)$ is the operator which returns the s.p.s.d. error matrix when optimally-truncating the argument down to rank $r$. Using $ \tilde {\mathcal M}_{B,i} = \rho \tilde {\mathcal M}_{B,i-1} + (1-\rho) M_i M_i^T$, equation (\ref{err_with_B_update_performed_2}) yields
\begin{equation}
E_i = \frac{1}{1-\rho}\text{Trc\_err}_{\text{SVD};r}\biggl(\rho \tilde {\mathcal M}_{B,i-1} + (1-\rho) M_i M_i^T\biggr)
\\\,\,\forall i\in\{1,...,k-1\}.
\label{err_with_B_update_performed_3}
\end{equation}
To proceed, we make the following two observations
\begin{enumerate}
	\item Since the SVD rank-$r$ truncation of $\rho \tilde {\mathcal M}_{B,i-1} + (1-\rho) M_i M_i^T$ is retaining the strongest $r$-modes, the error achieved by this truncation is minimal in $\norm{\cdot}_F$ across all possible rank-r matrices (SVD rank-$r$ truncation is optimal in unitary invariant norms) [12].
	
	\item By its definition, we have that $ \tilde {\mathcal M}_{B,i-1}$ is of rank $r$. Thus, the matrix $\rho \tilde {\mathcal M}_{B,i-1}$ in (\ref{err_with_B_update_performed_3}) is of rank $r$.
\end{enumerate}
Combining the two observations above (using $\rho \tilde {\mathcal M}_{B,i-1}$ as the suboptimal rank-$r$ truncation of $ \tilde {\mathcal M}_{B,i} = \rho \tilde {\mathcal M}_{B,i-1} + (1-\rho) M_i M_i^T$ in (\ref{err_with_B_update_performed_3})), we get that $\forall i\in\{1,2,...,k-1\}$,
\begin{equation}
\norm{ \text{Trc\_err}_{\text{SVD};r}\biggl(\rho \tilde {\mathcal M}_{B,i-1} + (1-\rho) M_i M_i^T\biggr) }_F \leq \norm{(1-\rho)M_iM_i^T}_F,
\label{eqn_gotta_explain_what_it_says}
\end{equation}
Substituting (\ref{eqn_gotta_explain_what_it_says}) in (\ref{err_with_B_update_performed_3}) after taking $\norm{\cdot}_F$ of the latter, gives
\begin{equation}
\norm{ E_i }_F \leq \norm{M_iM_i^T}_F, \,\,\,\forall i\in\{1,...,k-1\},
\label{eqn_prop_4_2_E_i_almost_done_B_updated}
\end{equation}
Note that equation (\ref{eqn_gotta_explain_what_it_says}) is essentially saying that the error of the SVD-based rank-$r$ truncation (l.h.s.) is at least as good as the error when choosing to represent the original matrix by the rank-$r$ matrix $\rho \tilde {\mathcal M}_{B,i-1}$ (which is of course a generally sub-optimal rank-$r$ truncation).

Trivially, \textit{Proposition 5.1} gives
\begin{equation}
E_k = 0,
\label{e_k_Is_zero_B_update_P_5_1_2}
\end{equation}
which together with (\ref{eqn_prop_4_2_E_i_almost_done_B_updated}) completes the proof for (\ref{eqn_2_to_prove_in_prop_4_2}) of \textit{Proposition 4.2}. $\square$
\section{SENG Hyperparameters used in Numerics}
\textbf{Hyper-parameters:} 

\noindent \textit{label\_smoothing = 0, fim\_col\_sample\_size = 128, lr\_scheme = 'exp', lr = 0.05, lr\_decay\_rate = 6, lr\_decay\_epoch = 75, damping = 2, weight\_decay = 1e-2, momentum = 0.9, curvature\_update\_freq = 200. Omitted params.\ are default.}
	